\documentclass[journal]{IEEEtran}
\usepackage[T1]{fontenc}
\usepackage{graphicx}
\usepackage{times}
\usepackage{helvet}
\usepackage{courier}
\usepackage{amsmath}
\usepackage{algorithm}
\usepackage{algorithmic}
\usepackage{csquotes} 
\usepackage{color}
\usepackage{paralist}
\usepackage{amssymb}
\usepackage{indentfirst}
\usepackage{subfigure}
\usepackage{float}
\usepackage{multirow}
\usepackage{cite}
\usepackage{mathrsfs}

\usepackage{mathrsfs} 
\usepackage{amsfonts}
\usepackage{todonotes}
\usepackage{pgfplots} 
\usepackage{epstopdf}
\usepackage{epsfig}
\pgfplotsset{compat=newest}
\usepackage{color}
\definecolor{forestgreen}{RGB}{0,139,69}

\usepackage{soul}
\usepackage{booktabs}
\usepackage{array}

\usepackage{xcolor}
\definecolor{citecolor}{HTML}{0071bc}
\usepackage[colorlinks, linkcolor=red,  anchorcolor=blue, citecolor=citecolor]{hyperref} 

\usepackage{xcolor}
\definecolor{SeaGreen4}{RGB}{0,205,102} 
\definecolor{SlateBlue}{RGB}{106,90,205} 
\definecolor{DarkRed}{RGB}{178,34,34} 
	
\usepackage[switch]{lineno}

\usepackage{textcomp,booktabs}
\usepackage{amssymb}
\usepackage{pifont}

\usepackage{makecell}

\usepackage{colortbl}
\definecolor{mygray}{gray}{.9}
\definecolor{mypink}{rgb}{.99,.91,.95}
\definecolor{mycyan}{cmyk}{.3,0,0,0}

\begin{document}

\title{ MAC-RRG: Iterative Multi-Agent Collaboration for X-ray Radiology Report Generation }

\author{ Futian Wang, Yuhan Qiao, Xiao Wang*, \emph{Member, IEEE}, Dan Xu, Yuehang Li, \\ 
            Zhixiang Guo*, Yaowei Wang, \emph{Member, IEEE}, Jin Tang

\thanks{ $\bullet$ Futian Wang, Yuhan Qiao, Xiao Wang, Dan Xu, Yuehang Li, and Jin Tang are with School of Computer Science and Technology, Anhui University, Hefei 230601, China. (email: \{wft, xiaowang, tangjin\}@ahu.edu.cn, \{e24301191, e23201112\}@stu.ahu.edu.cn, 18856245162@163.com)}

\thanks{ $\bullet$ Zhixiang Guo is with The First Affiliated Hospital of Anhui Medical University, Hefei 230000, China. 
(email: aydgzx100@163.com)} 

\thanks{$\bullet$ Yaowei Wang is with Harbin Institute of Technology, Shenzhen, China; Peng Cheng Laboratory, Shenzhen, China. (email: wangyw@pcl.ac.cn)} 

\thanks{* Corresponding Author: Xiao Wang $\&$ Zhixiang Guo}  
}

\markboth{ IEEE Transactions on ***, 2026 } 
{Shell \MakeLowercase{\textit{et al.}}: Bare Demo of IEEEtran.cls for IEEE Journals}

\maketitle



\begin{abstract}
Despite the remarkable progress of LLM-based and knowledge graph-augmented Radiology Report Generation (RRG) methods, existing techniques still suffer from inherent defects. Conventional LLM-only models lack structured medical prior knowledge, resulting in frequent medical hallucinations and low diagnostic interpretability. Current knowledge graph-enhanced schemes adopt static one-round knowledge fusion with single-source knowledge, incapable of dynamic knowledge updating according to generation feedback. This paper proposes a novel Multi-Agent Collaborative iterative framework for X-ray Radiology Report Generation, termed MAC-RRG. Inspired by multi-agent technology, our framework constructs a closed-loop optimization paradigm based on task decoupling and collaborative reasoning. Specifically, the framework first generates a preliminary radiology report from input X-ray images via a vision encoder and a basic LLM. Subsequently, a multimodal knowledge graph (MM-KG) agent mines structured disease correlation and anatomical knowledge from medical knowledge graphs, while an auxiliary knowledge agent extracts unstructured domain knowledge from public medical databases. The multi-source knowledge acquired by dual agents is fused and embedded to guide the LLM in iteratively refining the initial report. Extensive quantitative and qualitative experiments on mainstream X-ray RRG datasets, including IU X-ray, MIMIC, and CheXpert Plus, fully verify the superiority of our proposed method. The source code and pre-trained models have been released on \url{https://github.com/Event-AHU/Medical_Image_Analysis}
\end{abstract}

 
\IEEEpeerreviewmaketitle

\section{Introduction}

\IEEEPARstart{C}{hest} X-ray is one of the most widely adopted diagnostic imaging modalities in routine clinical screening, emergency diagnosis and respiratory disease surveillance. Radiology Report Generation (RRG)~\cite{messina2022survey}, which aims to automatically generate professional, accurate diagnostic descriptions corresponding to X-ray images, has attracted extensive research attention in the community of medical vision and large multimodal models. Manually writing radiology reports imposes a heavy workload on radiologists, often leading to prolonged waiting times for patients and increased risks of oversight or inconsistent descriptions caused by clinician fatigue. Automating X-ray RRG can effectively alleviate the labor burden of radiologists, streamline clinical workflows, and serve as an auxiliary diagnostic tool to improve the accessibility of radiological services, especially in regions facing shortages of professional radiology practitioners. Nevertheless, reliable X-ray RRG remains a challenging task: the model is required to precisely identify subtle radiographic lesions, capture interrelations among multiple imaging findings, and generate clinically consistent, logically rigorous diagnostic texts that conform to standardized radiological conventions~\cite{ganeshan2018structured}. 

\begin{figure*} 
\centering
\includegraphics[width=\linewidth]{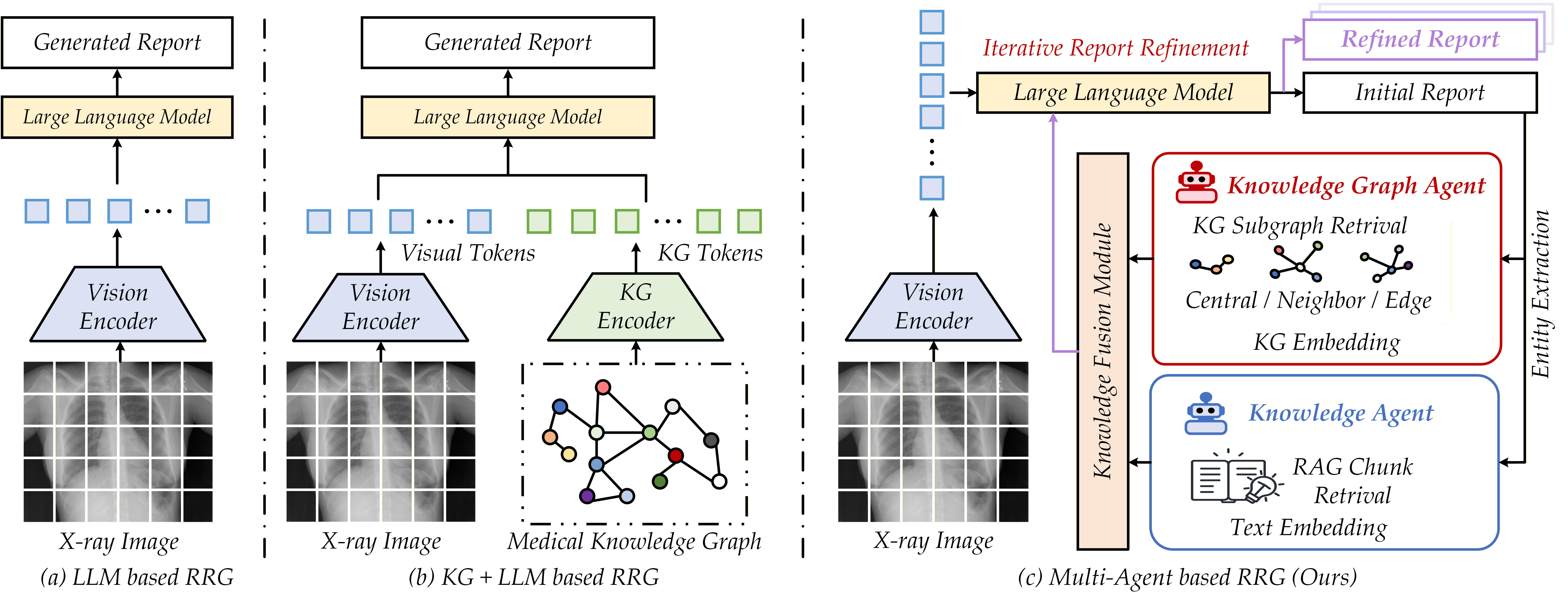}
\caption{Comparison between existing RRG frameworks and our newly proposed multi-agent RRG based model. (a) The baseline LLM-based medical report generation paradigm only fuses visual features extracted from raw X-ray images without incorporating structured prior medical knowledge. (b) Knowledge graph-enhanced LLM-based medical report generation models feed KG tokens encoded from medical knowledge graphs together with visual tokens into large language models. (c) We propose a novel iterative multi-agent collaboration framework for X-ray radiology report generation, termed MAC-RRG, which leverages collaborative multi-agent knowledge mining and fusion on initial radiology reports to iteratively enhance report quality. }
\label{fig:firstIMG}
\end{figure*}

In the deep learning era, the performance of X-ray RRG has been continuously advanced. 
Early research mainly built upon convolutional neural networks (CNNs)~\cite{lecun1998gradient}, such as the multi-task co-attention model proposed by Jing et al.~\cite{jing2018automatic}, TieNet~\cite{wang2018tienet}, which jointly performs thoracic disease classification and radiology report generation, and HRGR-Agent~\cite{li2018hybrid}, which combines template retrieval with free-text generation. These methods typically adopt an encoder–decoder paradigm, where CNNs extract visual features from X-ray images, followed by recurrent networks like LSTM~\cite{hochreiter1997long} to generate descriptive texts. Later, Transformer architectures~\cite{ashish2017attention} gradually became the mainstream backbone for RRG tasks, represented by R2Gen~\cite{chen2020generating}, which introduces a relational memory mechanism to enhance the modeling of report generation patterns and long-range dependencies in lengthy texts; R2GenCMN~\cite{chen2021cross} further employs a cross-modal memory network to explicitly model the correspondence between radiographic image regions and textual descriptions. Building upon these methods, PPKED~\cite{liu2021exploring} explores and distills prior and posterior knowledge to alleviate visual and linguistic biases caused by the imbalanced distribution of normal and abnormal cases; METransformer~\cite{wang2023metransformer} introduces multiple learnable expert tokens, enabling different experts to attend to complementary image regions and generate reports by aggregating their predictions.

Benefiting from the self-attention mechanism, Transformer-based models are capable of modeling long-range dependencies among imaging regions and output tokens, yielding clearer correspondence between visual lesions and textual descriptions. With the rapid advancement of pre-trained large multimodal models~\cite{li2023blip} in recent years, the quality of automatically generated X-ray radiology reports has achieved substantial improvements, as exemplified by R2GenGPT~\cite{wang2023r2gengpt}, which employs a lightweight visual alignment module to map radiographic image features into the word embedding space of a frozen large language model, thereby improving report generation capabilities while reducing the number of trainable parameters; R2GenCSR~\cite{wang2026r2gencsr}, which retrieves positive and negative contextual samples related to the current image to provide supplementary semantic references and jointly feeds visual features, retrieved contexts, and textual prompts into a large language model, thereby enhancing the discriminative capability of visual representations and improving report generation quality; and MambaXray-VL~\cite{wang2025cxpmrg}, which adopts a Mamba-based visual backbone and constructs a multi-stage pre-training framework integrating self-supervised autoregressive learning, image–report contrastive learning, and supervised fine-tuning to enhance the visual representations of X-ray images and cross-modal alignment capabilities. By leveraging large-scale image-text pre-training data, these models establish stronger cross-modal alignment between radiographic visual content and professional radiological language, further narrowing the performance gap between machine-generated narratives and manually written clinical reports.

Despite the aforementioned progress, we believe the following issues still limit these works: 
\textbf{(1).}  The baseline LLM-based medical report generation paradigm only fuses visual features extracted from raw X-ray images without incorporating structured prior medical knowledge, as shown in Fig.~\ref{fig:firstIMG} (a). As a result, the model cannot effectively leverage domain-specific medical knowledge such as disease correlations and anatomical priors. When interpreting complex imaging findings, it is prone to medical hallucinations, and its diagnostic reasoning lacks interpretability. 
\textbf{(2).} Knowledge graph-enhanced LLM-based medical report generation models feed KG tokens encoded from medical knowledge graphs together with visual tokens into large language models, as shown in Fig.~\ref{fig:firstIMG} (b), which mitigates the problem of insufficient external knowledge to a certain extent. However, the embedded knowledge graph features are integrated statically in a one-off manner during forward inference, making it impossible to dynamically adjust and iteratively retrieve required knowledge conditioned on preliminarily generated reports. Moreover, such methods only rely on structured graph knowledge from knowledge graphs, without supplementary unstructured textual knowledge including medical literature and clinical guidelines. 
\textbf{(3).} Both aforementioned approaches adopt a one-shot report generation pipeline without a feedback loop for iterative refinement. Errors, including entity omissions and biased descriptions of imaging findings in initial reports, cannot be continuously verified and corrected using external medical knowledge.

To tackle the inherent drawbacks of \textit{static knowledge fusion}, \textit{single-round generation}, and \textit{insufficient multi-source knowledge utilization} in existing RRG methods, we are inspired by the emerging multi-agent technology~\cite{hong2024metagpt,wu2023autogen}. Different from conventional static embedding and one-shot inference pipelines, the multi-agent collaboration mechanism possesses distinct advantages of task decoupling, specialized functional division, dynamic interactive reasoning, and iterative closed-loop optimization. Specifically, professionalized knowledge agents can be deployed to undertake targeted knowledge mining tasks separately, breaking the limitation of single structured knowledge sources. Meanwhile, the interactive iteration mechanism between agents and the generative LLM enables dynamic knowledge retrieval and real-time correction guided by initial report feedback, rather than passive one-time knowledge fusion. This hierarchical collaborative and iterative optimization paradigm perfectly compensates for the lack of interpretability, static knowledge fusion, and non-iterative generation in previous RRG frameworks, providing a feasible solution to the above-mentioned bottlenecks.

In this paper, we propose a novel iterative multi-agent collaboration framework for X-ray radiology report generation, termed MAC-RRG, which leverages collaborative multi-agent knowledge mining and fusion on initial radiology reports to iteratively enhance report quality. Given the X-ray image, we first partition it into multiple image patches and project them into visual tokens. A vision encoder is adopted to further enhance the features, and a Large Language Model (LLM) is used to produce an initial radiology report. Then, we extract the entities from the report and mine the key centrals/neighbors/edges from the multi-modal X-ray knowledge graph using the MM-KG agent, and extract relevant domain knowledge from existing public medical knowledge bases~\cite{chen2024towards} using another knowledge agent. Such knowledge is encoded into token embeddings and aggregated by the knowledge fusion module. The fused features are subsequently fed into the large language model to achieve iterative radiology report refinement. The comparison between existing radiology report generation models and our newly proposed framework can be found in Fig.~\ref{fig:firstIMG}. An overview of our MAC-RRG can be found in Fig.~\ref{fig:Retrieval_framework}.

To sum up, the main contributions of this paper can be summarized as follows: 

$\bullet$ We propose a novel iterative multi-agent collaboration framework for X-ray radiology report generation, termed MAC-RRG, which enables multiple agents to collaboratively perform supplementary knowledge mining and fusion over initial radiology reports for iterative report enhancement.

$\bullet$ We propose a multimodal knowledge graph agent, a knowledge agent, and a knowledge fusion module for X-ray radiology report generation, which jointly enable high-quality iterative refinement of radiology reports. 

$\bullet$ Extensive experiments on multiple mainstream X-ray radiology report datasets demonstrate the efficacy of our method. In comparison with conventional large-model approaches and knowledge graph-enhanced methods, the proposed model obtains improved interpretability and enhanced capabilities for disease diagnosis and analysis.

\textit{The rest of this paper is organized as follows:} We give a brief review of the most related works in Section~\ref{sec::relatedWorks}, and describe our framework in Section~\ref{sec::method}, with a focus on the overview of our framework, initial report generation and entity extraction, multi-agent based knowledge fusion, and the loss function. In Section~\ref{sec::experiments}, we conduct experiments to validate the effectiveness in both quality and quantity analysis. We conclude this paper in Section~\ref{sec::conclusion}.

\section{Related Works} \label{sec::relatedWorks}

\subsection{Radiology Report Generation} 
RRG aims to generate structurally complete and clinically factual free-text reports from medical images such as chest X-rays. Early studies typically followed the paradigm of image captioning by combining CNN-based visual encoders with recurrent decoders or hierarchical language models~\cite{jing2018automatic, wang2018tienet}, where co-attention mechanisms were employed to localize abnormal regions and generate multi-sentence reports. Subsequently, Transformer architectures and memory mechanisms have been widely adopted to model inter-sentence dependencies and the distribution of medical terminology in long-form reports. For example, R2Gen~\cite{chen2020generating} introduces a memory-driven Transformer to record key semantic information during the generation process, while R2GenCMN~\cite{chen2021cross} further employs cross-modal shared memory to explicitly model the alignment between visual and textual representations. These methods have substantially advanced end-to-end RRG; however, their language generation modules typically rely on training from scratch or large-scale fine-tuning on task-specific data. In long-text and fact-sensitive scenarios such as medical report generation, they remain prone to abnormality omission, vision–language mismatch, and template-like report generation.

In recent years, large language models (LLMs) have been introduced into RRG to leverage their strong language modeling ability, medical terminology organization, and instruction-following capability. R2GenGPT~\cite{wang2023r2gengpt} aligns visual features with the word embedding space of a frozen LLM, enabling efficient report generation by training only a lightweight alignment module. RaDialog~\cite{pellegrini2025radialog} further integrates visual features, structured pathological findings, and instruction data to support both report generation and interactive radiology-oriented dialogue. These LLM-based RRG methods demonstrate that LLMs can serve as powerful generators. Nevertheless, relying solely on visual alignment is still insufficient to ensure that each key entity is supported by adequate external medical evidence.

\subsection{Medical Knowledge Graphs}
Medical knowledge-enhanced methods focus on explicitly incorporating anatomical structures, disease concepts, and clinical relationships into radiology report generation models. KERP~\cite{li2019knowledge} decomposes medical image report generation into three stages, namely abnormality graph learning, retrieval, and paraphrasing, and provides early evidence for the value of explicit medical knowledge in improving report accuracy. PPKED~\cite{liu2021exploring} further explores posterior and prior knowledge from abnormal regions, medical knowledge graphs, and historical reports, and integrates such knowledge into report generation through knowledge distillation. KiUT~\cite{huang2023kiut} employs a symptom graph and a knowledge distiller to facilitate multi-level visual-textual interaction in a U-Transformer, while KARGEN~\cite{li2024kargen} introduces a disease-related knowledge graph within a frozen LLM framework to activate more relevant knowledge about thoracic diseases.

Structured radiology resources also provide an important foundation for knowledge-enhanced RRG. RadGraph~\cite{jain2021radgraph} annotates and extracts clinical entities and their relations from free-text radiology reports, while Chest ImaGenome~\cite{wu2021chest} represents chest X-rays as anatomy-centered scene graphs, indicating that clinical facts in radiology reports are naturally suited to entity-relation structured representations. Collectively, these studies suggest that report quality depends not only on whether a specific disease is correctly identified, but also on whether the relationships between diseases and anatomical locations, attributes, temporal changes, and adjacent findings are clinically consistent.

\subsection{Retrieval-Augmented Generation and Agentic Workflows}

Retrieval-Augmented Generation (RAG)~\cite{lewis2020retrieval} combines parametric generative models with non-parametric external memory, enabling models to generate more specific text by leveraging updatable and traceable external evidence. In chest X-ray report generation, Retrieval-Augmented Chest X-Ray Report Generation~\cite{ranjit2023retrieval} retrieves candidate report texts using multimodally aligned embeddings and feeds them into general-purpose GPT models for report generation. LaB-RAG~\cite{song2024lab} enhances textual retrieval with image-derived labels, while RA-RRG~\cite{park2026ra} employs clinical key phrase retrieval to reduce hallucinations and computational cost. These methods demonstrate that retrieval can compensate for medical knowledge missing or unstable in model parameters. However, most existing approaches rely on image-level or report-level similarity, and the retrieved results may not fully match the specific entities in the current sample, thereby introducing noisy evidence.

Agentic language model research further demonstrates that generation can be improved by decomposing complex tasks into intermediate reasoning, tool use, and knowledge acquisition steps. For example, ReAct~\cite{yao2022react} interleaves reasoning traces with task-specific actions, enabling language models to interact with external knowledge sources during inference. Inspired by this idea, we adopt a lightweight agentic workflow tailored to radiology report generation. Our framework instantiates role-specialized knowledge agents: an MM-KG Agent retrieves and aggregates structured entity relations from a medical knowledge graph, while a Knowledge Agent retrieves complementary textual evidence from an external medical knowledge corpus. The two knowledge streams are integrated through a knowledge fusion module to refine the initial report, enabling entity-aware, source-complementary, and clinically grounded report generation.

\section{Our Proposed Approach} \label{sec::method} 

\begin{figure*}[!htp]
    \centering
    \includegraphics[width=0.9\linewidth]{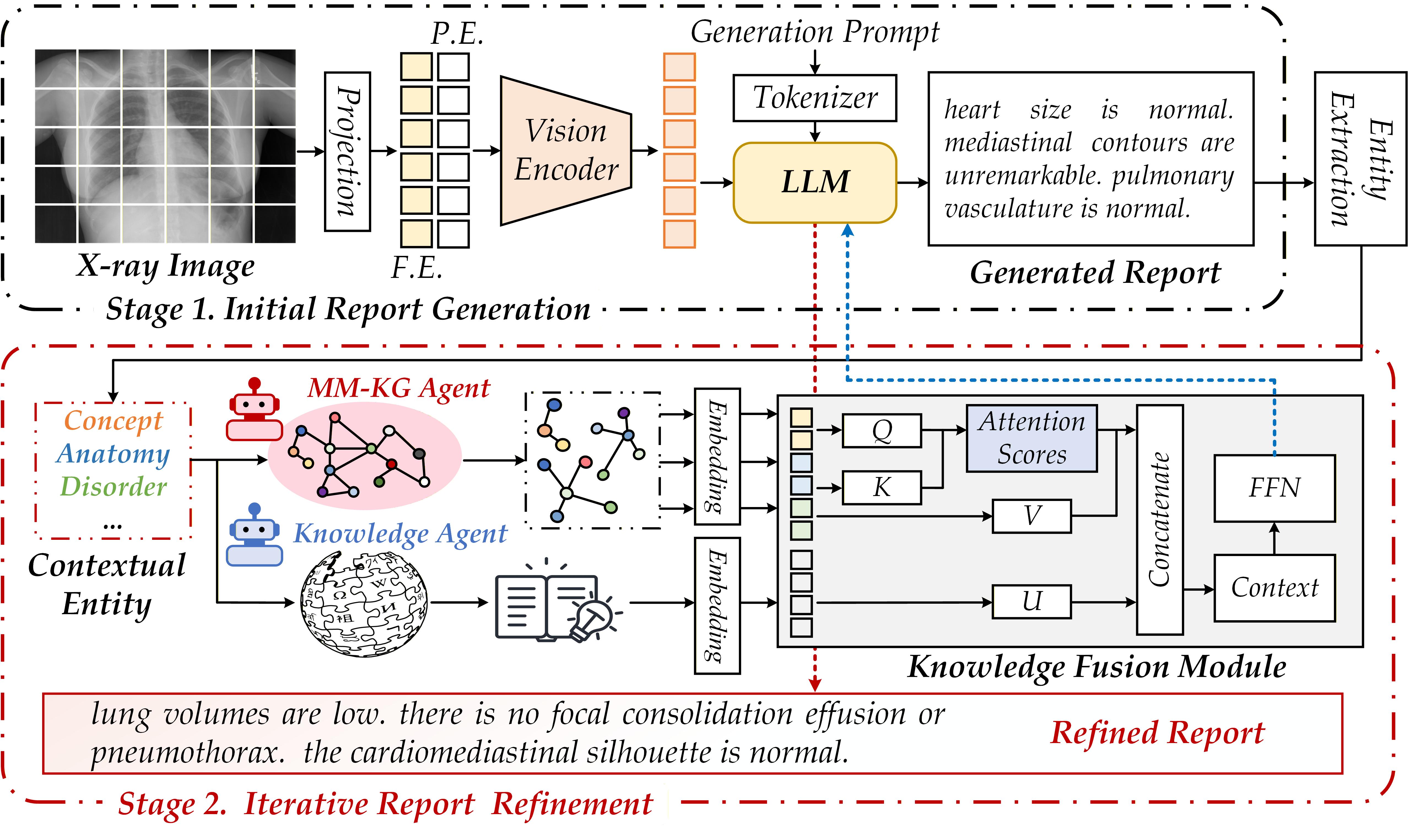}
    \caption{An overview of our iterative multi-agent collaboration for X-ray report generation framework, termed MAC-RRG. The framework comprises two stages: (1) Initial Report Generation, in which the X-ray image is encoded into visual tokens and fed into a large language model to generate an initial radiology report. (2) Iterative Report Refinement, where the extracted entities are used to construct local neighborhoods from a medical knowledge graph and retrieve relevant textual evidence. The graph knowledge and retrieval knowledge are encoded and fused with image tokens through the Knowledge Fusion Module, and the resulting multimodal representations are projected into the hidden space of the large language model to generate the final radiology report.}
    \label{fig:Retrieval_framework}
\end{figure*}


\subsection{Overview} 
As shown in Fig.~\ref{fig:Retrieval_framework}, we propose an iterative multi-agent collaboration-enhanced framework for X-ray radiology report generation, termed MAC-RRG. The framework first employs a visual encoder and a large language model to generate an initial report, from which anatomical structures and disease or abnormality concepts are subsequently extracted. These clinically meaningful entities are then processed through two complementary branches. In the graph branch, local neighborhoods are constructed based on a medical knowledge graph and encoded using Bio\_ClinicalBERT. In the retrieval branch, a BGE-based retriever and reranker are used to retrieve relevant textual evidence for each entity, which is likewise encoded using Bio\_ClinicalBERT. Finally, the graph knowledge tokens and retrieval knowledge tokens are integrated through the Knowledge Fusion Module, concatenated with the image tokens, and projected into the hidden space of the large language model. The decoder then generates radiology reports that are more consistent with established medical standards.

\subsection{Initial Report Generation and Entity Extraction}

The visual encoder maps chest X-ray images into a sequence of image tokens that preserve radiological evidence. Each input view $\mathbf{I}$ is encoded by the Swin Transformer \(f_v(\cdot)\) as follows:
\begin{equation}
    \mathbf{V} = f_v(\mathbf{I}) \in \mathbb{R}^{N_v \times d_v},
\end{equation}
where \(\mathbf{V}\) denotes the visual token sequence, \(N_v\) is the number of visual tokens, and \(d_v\) is the feature dimension. The visual sequence \(\mathbf{V}\) is then projected into the hidden dimension of the LLM and used to autoregressively generate an initial report.

The initial report provides explicit textual anchors for selecting clinically relevant knowledge. As illustrated in the upper panel of Fig.~\ref{fig:kg_chunks}, the draft report \(R_0\) serves as the starting point for subsequent knowledge retrieval. Given the draft report \(R_0\), the entity extraction module first segments it into sentences, normalizes punctuation and letter case, and performs lexical matching based on a radiology entity dictionary. We retain only two types of entities, namely \emph{anatomy} and \emph{disorder}, since they directly describe the most essential anatomical regions and abnormal findings in chest X-ray reports. Duplicate entity mentions across sentences are then merged to form an examination-level entity set:
\begin{equation}
    \mathcal{E} = \{(e_i, t_i)\}_{i=1}^{M},
\end{equation}
where \(e_i\) denotes the normalized entity mention, and \(t_i \in \{\mathrm{anatomy}, \mathrm{disorder}\}\) denotes the entity type.

Using the initial report for entity selection has two advantages. First, it restricts external knowledge queries to image-conditioned clinical concepts, rather than retrieving general radiological priors. Second, it provides a shared entity set for both the graph branch and the retrieval branch, enabling structured and unstructured knowledge to remain comparable at the token level.

\subsection{MM-KG Agent}

The MM-KG Agent injects structured radiological relations into the report generation process by expanding each preliminary report entity into a local knowledge graph domain. The medical knowledge graph is stored as a collection of weighted triples. For each entity $e_i$, the agent searches for triples whose source or target entity matches $e_i$. The matched edges are ranked according to frequency counts, and the top-$K_g$ neighboring entities are retained:
\begin{equation}
\mathcal{G}_i = \left\{ (e_i, r_{ij}, n_{ij}) \right\}_{j=1}^{K_i},
\end{equation}
where $n_{ij}$ denotes the neighboring entity and $r_{ij}$ represents the corresponding relation type. In the current setting, each entity is linked to at most ten neighboring entities, i.e., $K_g = 10$.

The central entity, neighboring entities, and relation labels are encoded using a frozen Bio\_ClinicalBERT encoder. Formally, we define
\begin{equation}
\hat{\mathcal{G}}_i = \left\{ (\phi(e_i), \phi(n_{ij}), \phi(r_{ij})) \right\}_{j=1}^{K_i},
\end{equation}
where $\phi(\cdot)$ denotes the Bio\_ClinicalBERT text encoding function. These embeddings are subsequently fed into the Knowledge Fusion Module, where they are fused with the retrieval-based knowledge representations produced by the Knowledge Agent.

This graph encoding design aims to preserve the relational structure among anatomical regions, imaging findings, and diseases. Rather than processing each concept independently, the model is provided with local graph context, enabling it to capture clinically meaningful associations during report generation.

\subsection{Knowledge Agent}

The Knowledge Agent complements the graph branch with unstructured medical textual evidence. Specifically, we construct a retrieval corpus based on the PubMedVision dataset by extracting visual question-answering samples related to chest X-rays (CXR). Each question-answer pair is then summarized by a large language model into a structured knowledge item. These knowledge items are sequentially collected as knowledge snippets for subsequent retrieval.

Different from conventional similar-report retrieval methods, the external knowledge base used in this work is collected and organized from online medical resources, and is further summarized by a large language model into compact entity-level knowledge entries. This design offers two advantages. First, the retrieval query is explicitly grounded in the medical entities appearing in the current report, which reduces irrelevant evidence introduced by global case-level similarity or long-text retrieval. Second, compared with raw web text or complete reports, the retrieved knowledge entries are more focused and contain less redundancy, thereby providing clearer clinical evidence for key medical entities.

For each extracted entity $e_i$, the retrieval query is constructed from the complete preliminary report, the target entity, and its entity type:
\begin{equation}
\mathbf{q}_i = [R_0; e_i; t_i].
\end{equation}

The entity-wise retriever first encodes $\mathbf{q}_i$ using BGE-M3 and retrieves the top-$N$ candidate chunks from a cached vector database according to inner-product similarity. The candidate chunks are then re-ranked by a BGE reranker, and the top-$K_r$ results are retained for each entity. In the current experiments, we set $N=10$ and retain one re-ranked chunk for each entity.

Since different entities may retrieve the same supporting text, the retrieved chunks are merged and deduplicated according to their original chunk indices. Let:
\begin{equation}
\mathcal{C} = \{c_1, \ldots, c_{N_r}\},
\end{equation}
denote the deduplicated evidence list. Each chunk is encoded by a frozen Bio\_ClinicalBERT encoder with mean pooling:
\begin{equation}
\mathbf{r}_j =
\operatorname{Normalize}
\left(
\frac{\sum_{\ell} a_{j}^{\ell} \mathbf{H}_{j}^{\ell}}
{\sum_{\ell} a_{j}^{\ell}}
\right),
\end{equation}
where $\mathbf{H}_{j}^{\ell}$ denotes the hidden state of the $\ell$-th token in the $j$-th chunk $c_j$, and $a_{j}^{\ell}$ denotes the corresponding attention mask. The retrieved tokens are then stacked and padded within each batch:
\begin{equation}
\mathbf{R} = [\mathbf{r}_1; \ldots; \mathbf{r}_{N_r}]
\in \mathbb{R}^{N_r \times 768}.
\end{equation}

The retrieval branch provides descriptive textual evidence that may not be explicitly represented in the triples of the graph branch. Encoding the retrieved chunks into continuous vectors avoids directly concatenating long textual prompts, thereby preserving entity-relevant external knowledge while keeping the input to the LLM compact.

\begin{figure*}[!htp]
\centering
\includegraphics[width=\linewidth]{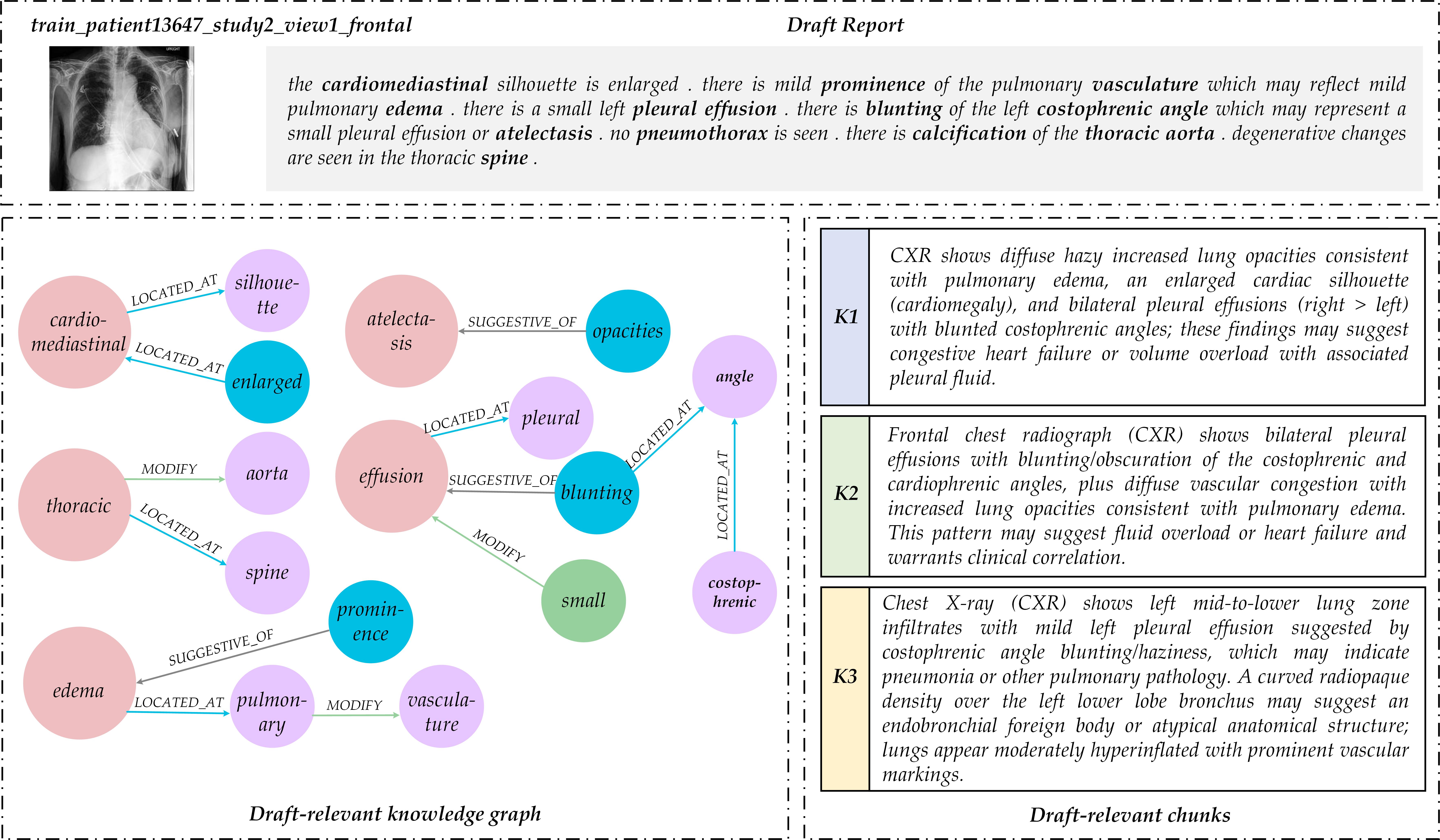}
\caption{The upper panel shows the input chest X-ray image and its corresponding Draft Report. The lower-left panel presents representative Draft-relevant knowledge graph
triples retrieved by the MM-KG Agent and organized into five independent entity-centered subgraphs. The lower-right panel shows representative Draft-relevant text chunks retrieved by the Knowledge Agent. The two agents provide complementary structured and descriptive evidence for subsequent knowledge fusion and report refinement. }
\label{fig:kg_chunks}
\end{figure*}

\subsection{knowledge Fusion Module}

The graph branch and the retrieval branch provide complementary knowledge. However, directly feeding such knowledge into the large language model would result in an excessively long prompt with substantial redundancy, which may limit its effectiveness in enhancing the model. Therefore, we introduce a knowledge fusion module.

We first fuse the knowledge from the graph branch. Specifically, the previously obtained embeddings \(h_i\), \(h_{ij}\), and \(q_{ij}\) are fed into a relation-aware graph attention encoder. For each attention head, the central entity embedding is transformed into a query, while the neighbor key is obtained by adding the corresponding edge embedding:
\[
Q_i = W_Q h_i, \quad
K_{ij} = W_K h_{ij} + q_{ij}, \quad
U_{ij} = W_V h_{ij}.
\]

The attention weight assigned to the \(j\)-th neighbor is computed as:
\[
\alpha_{ij}
=
\frac{
\exp\left( Q_i K_{ij}^{\top} / \sqrt{d} \right)
}{
\sum_{\ell=1}^{K_i}
\exp\left( Q_i K_{i\ell}^{\top} / \sqrt{d} \right)
}.
\]

The graph contextual representation of entity \(e_i\) is then obtained by:
\[
g_i =
\mathrm{MLP}
\left(
W_O
\sum_{j=1}^{K_i}
\alpha_{ij} U_{ij}
\right).
\]

The outputs of all entities are concatenated and dynamically padded to form the graph knowledge tokens:
\[
G = [g_1; \ldots; g_M] \in \mathbb{R}^{N_g \times 768}.
\]

For the retrieval branch, as described above, the retrieved results are first merged and deduplicated. Each resulting chunk is then encoded using Bio-ClinicalBERT to obtain the retrieval-based knowledge representation \(R\).

Finally, to fuse the knowledge from the graph and retrieval branches, we first project both representations into the same dimensional space as the visual features:
\[
\tilde{G} = W_g G, \quad \tilde{R} = W_r R.
\]

Let the image tokens output by the visual encoder be denoted as
\(V \in \mathbb{R}^{N_v \times d_v}\). The final multi-source knowledge representation is obtained through direct concatenation:
\[
Z = [V; \tilde{R}; \tilde{G}].
\]

The fused representation $Z$ is projected into the hidden space of the large language model and normalized:
\[
E = \mathrm{LayerNorm}(W_{\mathrm{llm}} Z).
\]

Finally, \(E\), together with the textual generation prompt, is used as the input prefix of the large language model for radiology report generation.



\subsection{Loss Function}

MAC-RRG is optimized using an autoregressive language modeling
objective. After integrating the visual tokens, graph knowledge
tokens, and retrieval knowledge tokens, the fused multimodal
representation $E$ is combined with the textual generation prompt
embedding $P$ to construct the conditioning prefix:
\begin{equation}
H = \operatorname{Concat}(E, P).
\end{equation}

Let $Y=(y_1,y_2,\ldots,y_T)$ denote the ground-truth radiology
report, where $T$ is the number of report tokens. Conditioned on
the multimodal prefix $H$, the probability of the target report is
factorized in an autoregressive manner as
\begin{equation}
p_{\theta}(Y \mid H)
=
\prod_{t=1}^{T}
p_{\theta}
\left(
y_t \mid y_{<t}, H
\right),
\end{equation}
where $y_{<t}$ denotes the report tokens preceding $y_t$, and
$\theta$ represents the trainable model parameters.

The language modeling objective is defined as the average negative
log-likelihood over the target report tokens:
\begin{equation}
\mathcal{L}_{\mathrm{LM}}
=
-\frac{1}{T}
\sum_{t=1}^{T}
\log
p_{\theta}
\left(
y_t \mid y_{<t}, H
\right).
\end{equation}

During training, the labels corresponding to the textual prompt,
visual tokens, graph knowledge tokens, and retrieval knowledge
tokens are masked with $-100$, such that only the ground-truth
radiology report tokens contribute to the language modeling loss.

\section{Experiments} \label{sec::experiments} 

\subsection{Datasets and Evaluation Metric} 

\begin{table*}[]
\caption{Comparison of our model’s performance on IU X-ray and Chexpert plus datasets. The best result is highlighted in bold.}
\label{tab:results_iu_chexpertplus}
\resizebox{\linewidth}{!}{
\begin{tabular}{c|l|l|ccccccc}
\hline \toprule [0.5 pt] 
\textbf{Dataset} & \textbf{Methods} & \textbf{Publication} & \textbf{BLEU-1} & \textbf{BLEU-2} & \textbf{BLEU-3} & \textbf{BLEU-4} & \textbf{ROUGE-L} & \textbf{METEOR} & \textbf{CIDEr} \\ 
 \hline \toprule [0.5 pt] 
 \multirow{13}{*}{\textbf{IU X-Ray}} 
 & R2Gen~\cite{chen2020generating} & EMNLP 2020 & 0.470 & 0.304 & 0.219 & 0.165 & 0.371 & 0.187 & - \\
& SentSAT+KG~\cite{zhang2020radiology} & AAAI 2020 & 0.441 & 0.291 & 0.203 & 0.147 & 0.367 & - & - \\
 & R2GenCMN~\cite{chen2021cross} & ACL-IJCNLP 2021 & 0.475 & 0.309 & 0.222 & 0.170 & 0.375 & 0.191 & - \\
 & PPKED~\cite{liu2021exploring} & CVPR 2021 & 0.483 & 0.315 & 0.224 & 0.168 & 0.376 & 0.187 & 0.351 \\
 & AlignTrans~\cite{you2021aligntransformer} & MICCAI 2021 & 0.484 & 0.313 & 0.225 & 0.173 & 0.379 & 0.204 & - \\
 & CMCL~\cite{liu2021competence} & ACL 2021 &0.473 & 0.305 & 0.217 & 0.162 & 0.378 & 0.186 & - \\
 & DCL~\cite{li2023DCL} & CVPR 2023 & - & - & - & 0.163 & 0.383 & 0.193 & 0.586 \\
 & R2GenGPT~\cite{wang2023r2gengpt} & Meta Radiology 2023 & 0.465 & 0.299 & 0.214 & 0.161 & 0.376 & 0.219 & 0.542 \\
 & PromptMRG~\cite{jin2024promptmrg} & AAAI 2024 & 0.401 & - & - & 0.098 & 0.160 & \textbf{0.281} & - \\ 
 & SILC~\cite{liu2024multi} & IEEE TMI 2024 & 0.472 & 0.321 & 0.234 & 0.175 & 0.379 & 0.192 & 0.368 \\
 & MMR ~\cite{fang2025automated} & BSPC 2025 & 0.497  & \textbf{0.333} & 0.240 & 0.185 & 0.399 & 0.215 & - \\
 & DuCo-Net~\cite{rahman2025duco} & IEEE Access 2025 & \textbf{0.500} & 0.330 & 0.220 & 0.160 & 0.260 & 0.240 & - \\
 & DVAF-DMSR~\cite{wu2026disease} & AAAI 2026 & 0.495 & 0.328 & 0.242 & 0.187 & 0.384 & - & 0.634 \\
& DVPAlign~\cite{sun2026dvpalign} &  ICASSP 2026 & 0.431 & - & - & 0.110 & 0.318 & 0.166 & - \\

 \cline{2-10} 

 & MAC-RRG & Ours & 0.488 & 0.330 & \textbf{0.249} & \textbf{0.197} & \textbf{0.400} & 0.223 & \textbf{0.703} \\
 \hline \toprule [0.5 pt] 
\multirow{13}{*}{\textbf{CheXpert Plus}} 
 & R2Gen~\cite{chen2020generating} & EMNLP 2020 & 0.301 & 0.179 & 0.118 & 0.081 & 0.246 & 0.113 & 0.077 \\
 & R2GenCMN~\cite{chen2021cross} & ACL-IJCNLP 2021 & 0.321 & 0.195 & 0.128 & 0.087 & 0.256 & 0.127 & 0.102 \\
 & XProNet~\cite{wang2022cross} & ECCV 2022 & 0.364 & 0.225 & 0.148 & 0.100 & 0.265 & 0.146 & 0.121 \\
 & ORGan~\cite{hou2023organ} & ACL 2023 & 0.320 & 0.196 & 0.128 & 0.086 & 0.261 & 0.135 & 0.107 \\
 & R2GenGPT~\cite{wang2023r2gengpt} & Meta Radiology 2023 & 0.361 & 0.224 & 0.149 & 0.101 & 0.266 & 0.145 & \textbf{0.123} \\
 & ASGMD~\cite{xue2024generating} & ESWA 2024 & 0.267 & 0.149 & 0.094 & 0.063 & 0.220 & 0.094 & 0.044 \\
 & Token-Mixer~\cite{yang2024token} & IEEE TMI 2024 & \textbf{0.378} & 0.231 & 0.153 & 0.091 & 0.262 & 0.135 & 0.098 \\
 & PromptMRG~\cite{jin2024promptmrg} & AAAI 2024 & 0.326 & 0.174 & - & 0.095 & 0.222 & 0.121 & 0.044 \\
 & MCA-RG~\cite{xing2025mca} & MICCAI 2025 & 0.367 & 0.218 & 0.149 & 0.102 & 0.266 & 0.147 & - \\
 & VLCI~\cite{chen2025cross} &  IEEE TIP 2025 & - & - & - & 0.080 & 0.247 & - & 0.072 \\
 & DVPAlign~\cite{sun2026dvpalign} &  ICASSP 2026 & 0.335 & - & - & 0.093 & 0.228 & 0.127 & - \\
 \cline{2-10} 
 & MAC-RRG &Ours & 0.375  & \textbf{0.232} & \textbf{0.154} & \textbf{0.105} & \textbf{0.267} & \textbf{0.151} & 0.121 \\
 \hline \toprule [0.5 pt] 
\end{tabular}
}
\end{table*}

\begin{table*}[]
\caption{Comparison of our model’s performance on the MIMIC-CXR datasets. The best result is highlighted in bold.}
\label{tab:results_mimic}
\resizebox{\linewidth}{!}{
\begin{tabular}{c|l|l|ccccccc}
\hline \toprule [0.5 pt] 
\textbf{Dataset} & \textbf{Methods} & \textbf{Publication} & \textbf{BLEU-1} & \textbf{BLEU-2} & \textbf{BLEU-3} & \textbf{BLEU-4} & \textbf{ROUGE-L} & \textbf{METEOR} & \textbf{CIDEr} \\ \hline
\multirow{13}{*}{\textbf{MIMIC-CXR}} 
 & R2Gen~\cite{chen2020generating} & EMNLP 2020 & 0.353 & 0.218 & 0.145 & 0.103 & 0.277 & 0.142 & - \\
 & R2GenCMN~\cite{chen2021cross} & ACL-IJCNLP 2021 & 0.353 & 0.218 & 0.148 & 0.106 & 0.278 & 0.142 & - \\
 & PPKED~\cite{liu2021exploring} & CVPR 2021 & 0.360 & 0.224 & 0.149 & 0.106 & 0.284 & 0.149 & 0.237 \\
 & AlignTrans~\cite{you2021aligntransformer} & MICCAI 2021 & 0.378 & 0.235 & 0.156 & 0.112 & 0.283 & 0.158 & - \\
 & CMCL~\cite{liu2021competence} & ACL 2021 &0.344 & 0.217 & 0.140 & 0.097 & 0.281 & 0.133 & - \\
 & Clinical-BERT~\cite{clinicalBert} & AAAI 2022 & 0.383 & 0.230 & 0.151 & 0.106 & 0.275 & 0.144 & 0.151 \\
 & METransformer~\cite{wang2023metransformer} & CVPR 2023 & 0.386 & 0.250 & 0.169 & 0.124 & \textbf{0.291} & 0.152 & \textbf{0.362} \\
 & DCL~\cite{li2023DCL} & CVPR 2023 & - & - & - & 0.109 & 0.284 & 0.150 & 0.281 \\
 & R2GenGPT~\cite{wang2023r2gengpt} & Meta Radiology 2023 & 0.405 & 0.252 & 0.171 & 0.123 & 0.285 & 0.167 & 0.254 \\
 & PromptMRG~\cite{jin2024promptmrg} & AAAI 2024 & 0.398 & - & - & 0.112 & 0.268 & 0.157 & - \\ 
 & AdaMatch-Cyclic~\cite{chen2024AdaMatch_Cyclic} & ACL 2024 & 0.379 & 0.235 & 0.154 & 0.106 & 0.286 & 0.163 & - \\ 
 & DACG~\cite{lang2025dacg} & MIA 2025 & 0.398 & 0.249 & 0.167 & 0.117 & 0.290 & 0.162 & - \\ 
 & Teaser~\cite{10770258} &  IEEE TMI 2025 & \textbf{0.423} & 0.257 & 0.166 & 0.113 & 0.287 & \textbf{0.170} & -
  \\
 & GDMRG~\cite{tang2026graph} & arXiv 2026 & 0.406 & - & - &  0.114 & 0.272 & 0.157 & -
 \\ 
 & DVPAlign~\cite{sun2026dvpalign} &  ICASSP 2026 & 0.408 & - & - & 0.118 & 0.276 & 0.164 & - \\
 \cline{2-10} 
 & MAC-RRG & Ours & 0.409 & \textbf{0.258} & \textbf{0.177} & \textbf{0.128} & 0.285 & 0.165 & 0.248 \\ 
\hline  \toprule [0.5 pt] 
\end{tabular}
}
\end{table*}

In our experiments, we employ three benchmark datasets that are widely used in the field of chest X-ray report generation, namely IU-Xray~\cite{demner2016iuxray}, MIMIC-CXR~\cite{johnson2019mimicCXR}, and CheXpert Plus~\cite{chambon2024CheXpertPLUS}, to evaluate the effectiveness of the proposed MAC-RRG model. The linguistic quality of the generated reports is assessed using commonly adopted natural language generation (NLG) metrics, including BLEU~\cite{papineni2002bleu}, ROUGE-L~\cite{lin2004rouge}, METEOR~\cite{banerjee2005meteor}, and CIDEr~\cite{vedantam2015cider}. In addition, following the evaluation protocol of R2Gen~\cite{chen2020generating}, we employ Clinical Efficacy (CE) metrics to evaluate the clinical accuracy of the generated reports. Detailed descriptions of the datasets and evaluation metrics are provided below.

\noindent $\bullet$ \textbf{IU-Xray Dataset.}
IU-Xray is a publicly available medical imaging dataset consisting primarily of chest X-ray images and their corresponding radiology reports. Released by the Indiana University School of Medicine, the dataset contains 7,470 chest X-ray images and 3,955 associated radiology reports. To ensure comparability with existing methods, we follow the experimental settings adopted by R2Gen~\cite{chen2020generating} and R2GenGPT~\cite{wang2023r2gengpt}, and partition the dataset into training, testing, and validation sets at a ratio of 7:1:2.

\noindent $\bullet$ \textbf{CheXpert Plus Dataset.}
CheXpert Plus is a large-scale multimodal dataset comprising both medical images and textual radiology reports. It was developed to improve the performance, robustness, and fairness of machine learning models in radiology. The dataset contains 223,228 chest X-ray images together with their corresponding reports and provides annotations for 14 categories of thoracic abnormalities, thereby offering high-quality data for research on medical image analysis and radiology report generation. CheXpert Plus has been applied to a variety of tasks, including disease diagnosis, medical image classification, abnormality labeling, and radiology report generation. To ensure fair and reproducible evaluation, we adopt the dataset partitioning protocol proposed in CXPMRG-Bench~\cite{wang2025cxpmrg}.

\noindent $\bullet$ \textbf{MIMIC-CXR Dataset.}
MIMIC-CXR (Medical Information Mart for Intensive Care Chest X-ray) is a large-scale chest X-ray dataset jointly developed by the Massachusetts Institute of Technology and Beth Israel Deaconess Medical Center. It contains 377,110 chest X-ray images and 227,835 radiology reports. To enable a fair comparison with existing radiology report generation methods, we follow the experimental settings of R2Gen~\cite{chen2020generating} and R2GenGPT~\cite{wang2023r2gengpt}, and use the official MIMIC-CXR splits for model training, validation, and testing.

For the evaluation of linguistic generation quality, BLEU measures the textual similarity between generated and reference reports based on $n$-gram precision. ROUGE-L evaluates content overlap by computing the longest common subsequence (LCS) between a generated report and its reference report. METEOR extends lexical matching by accounting for morphological variations, synonymous expressions, and differences in word order, thereby alleviating some of the limitations of BLEU in evaluating semantic consistency. CIDEr employs TF--IDF-weighted $n$-gram matching to emphasize informative textual expressions while reducing the excessive influence of frequently occurring phrases on the evaluation results.

In addition to linguistic evaluation, we employ Precision, Recall, and F1 score to assess the ability of MAC-RRG to recognize diseases, lesions, and other clinically relevant abnormalities. Precision measures the proportion of correctly identified positive clinical labels among all labels predicted as positive; therefore, a higher precision indicates fewer false-positive predictions. Recall measures the proportion of actual positive clinical labels that are successfully identified by the model, with a higher recall indicating that the model captures a larger proportion of clinically relevant abnormalities. The F1 score is the harmonic mean of Precision and Recall and thus provides a comprehensive measure of the model's overall performance in clinical abnormality recognition.

\begin{table}
\centering
\small 
\caption{A comparison of the clinical efficacy (CE) metrics between our proposed framework (Ours) and state-of-the-art methods using F1 score, precision, and recall on the CheXpert Plus dataset.}
\label{tab:ce_metrics}
\begin{tabular}{l|ccc}
\hline \toprule [0.5 pt] 
\textbf{Model} & \textbf{F1} & \textbf{Precision} & \textbf{Recall} \\
\hline
R2Gen~\cite{chen2020generating} & 0.181 & 0.318 & 0.200 \\
R2GenCMN~\cite{chen2021cross} & 0.231 & 0.329 & 0.241 \\
WCL~\cite{yan2021weakly}& 0.256 & 0.335 & 0.259 \\
PromptMRG~\cite{jin2024promptmrg} & 0.281 & 0.258 & 0.265 \\
R2GenGPT ~\cite{wang2023r2gengpt} & 0.260 & 0.315 & 0.224 \\
ORGan~\cite{hou2023organ} & 0.277 & 0.288 & \textbf{0.287} \\
Token-Mixer~\cite{yang2024token} & 0.288 & 0.309 & 0.270 \\
VLCI~\cite{chen2025cross} & 0.163 &  0.341 &  0.175 \\
\cline{1-4} 
MAC-RRG (Ours) & \textbf{0.295} & \textbf{0.350} & 0.274 \\
\hline \toprule [0.5 pt] 
\end{tabular}
\end{table}

\begin{table*}[ht]
\centering
\small 
\caption{Ablation study on Chexpert plus dataset, assessing the impact of key components: MM-KG Agent (MM-KG), Knowledge Agent (KA), knowledge Fusion Module (KFM). A “\checkmark” indicates the presence of each component, while “-” denotes its absence.}
\label{tab:ablation_nlg_chexpertplus}
\resizebox{\textwidth}{!}{
\begin{tabular}{ c | c | c c c | c c c c c c c c c c }
\hline \toprule [0.5 pt] 
\textbf{Dataset} & \textbf{Setting} & \textbf{MM-KG} & \textbf{KA} & \textbf{KFM}   
& \textbf{BLEU-1} & \textbf{BLEU-2} & \textbf{BLEU-3} & \textbf{BLEU-4} 
& \textbf{RG-L} & \textbf{METEOR} & \textbf{CIDEr} & \textbf{F1} & \textbf{Precision} & \textbf{Recall} \\
\hline
\multirow{5}{*}{CheXpert Plus} 
& BASE & - & - & - & 0.361 & 0.224 & 0.149 & 0.101 & 0.266 & 0.145 & 0.123 & 0.260 & 0.315 & 0.224\\
& (a)  & \checkmark & - & - &  0.366  & 0.227 & 0.150 & 0.103 & 0.266 & 0.148 & 0.124 & 0.286 & 0.340 & 0.270\\
& (b)  & - & \checkmark & - & 0.370 & 0.228 & 0.151 & 0.103 & 0.266 & 0.149 & 0.125 & 0.288 & 0.344 & 0.269\\
& (c)  & \checkmark & \checkmark & \- & 0.372 & 0.231 & 0.153 & 0.104 & 0.265 & 0.150 & 0.122 & 0.292 & 0.359 & 0.270 \\
\cline{2-15}
& (d)  & \checkmark & \checkmark & \checkmark & 0.375  & 0.232 & 0.154 & 0.105 & 0.267 & 0.151 & 0.121 & 0.295 & 0.350 & 0.274\\
\hline \toprule [0.5 pt] 
\end{tabular}
} 
\end{table*}

\subsection{Implementation Details} 
We use a pre-trained Swin Transformer~\cite{liu2021swin} as the visual encoder and Llama2-7B~\cite{touvron2023llama} as the language backbone. Contextual entities extracted from the initial draft report are used to query both the medical knowledge graph and the external chunk memory. For the MM-KG Agent, each entity retrieves up to 10 KG neighboring nodes, and the center nodes, neighboring nodes, and relation edges are encoded into 768-dimensional features using frozen Bio ClinicalBERT~\cite{alsentzer2019publicly}. For the Knowledge Agent, each entity retrieves candidate chunks from a chunk memory containing 393 textual knowledge entries, followed by reranking. Then, the KG features and retrieval features are fused and mapped to the 1024-dimensional visual feature space. They are concatenated with visual tokens and further projected into the 4096-dimensional LLM embedding space for report generation. The model is implemented in PyTorch~\cite{paszke2019pytorch} and optimized with AdamW~\cite{loshchilov2017decoupled} using a learning rate of \(1\times10^{-4}\). A cosine annealing scheduler is adopted with a minimum learning rate of \(1\times10^{-6}\). All experiments are conducted on an NVIDIA A800SXM4-80GB GPU. More details can be found in our source code.

\subsection{Comparison on Public Benchmark Datasets} 

\noindent $\bullet$ \textbf{Analysis of the NLG Metrics.~} To comprehensively evaluate the effectiveness of the proposed method for medical image report generation, we conduct comparative experiments on three widely used public benchmark datasets, including IU X-Ray, CheXpert Plus, and MIMIC-CXR. We adopt commonly used natural language generation evaluation metrics, including BLEU, ROUGE-L, METEOR, and CIDEr. The compared methods cover traditional encoder-decoder frameworks, cross-modal alignment methods, pre-trained language model enhanced methods, and knowledge-/graph-enhanced methods, thereby providing a comprehensive evaluation of the report generation capability of the proposed method on datasets with different scales and complexities.

As shown in Table~\ref{tab:results_iu_chexpertplus}, on the IU X-Ray dataset, the proposed MAC-RRG achieves competitive overall performance, with BLEU-1, BLEU-2, BLEU-3, BLEU-4, ROUGE-L, METEOR, and CIDEr scores of 0.488, 0.330, 0.249, 0.197, 0.400, 0.223, and 0.703, respectively. Specifically, MAC-RRG achieves the best results on BLEU-3, BLEU-4, ROUGE-L, and CIDEr. This indicates that the proposed method not only maintains good word-level matching capability, but also has stronger ability in generating continuous medical phrases, complex semantic fragments, and complete report contents.

On the more complex CheXpert Plus dataset, MAC-RRG also demonstrates stable and excellent generation performance. As shown in Table~\ref{tab:results_iu_chexpertplus}, the proposed method achieves BLEU-1, BLEU-2, BLEU-3, BLEU-4, ROUGE-L, METEOR, and CIDEr scores of 0.375, 0.232, 0.154, 0.105, 0.267, 0.151, and 0.121, respectively. Among these metrics, MAC-RRG achieves the best performance on BLEU-2, BLEU-3, BLEU-4, ROUGE-L, and METEOR, indicating that the proposed method can better capture semantic dependencies among medical entities and generate medical descriptions that are highly consistent with the reference reports. Although the BLEU-1 score is slightly lower than that of Token-Mixer, which achieves 0.378, and the CIDEr score is slightly lower than that of R2GenGPT, which achieves 0.123, the gaps are relatively small. This suggests that the proposed method maintains strong text matching capability while paying more attention to the consistency of medical semantic relations, contextual structure, and overall report quality.

On the larger-scale MIMIC-CXR dataset with more complex clinical descriptions, MAC-RRG further demonstrates good generalization ability and stability. As shown in Table~\ref{tab:results_mimic}, the proposed method achieves BLEU-1, BLEU-2, BLEU-3, BLEU-4, ROUGE-L, METEOR, and CIDEr scores of 0.409, 0.258, 0.177, 0.128, 0.285, 0.165, and 0.248, respectively. Specifically, MAC-RRG obtains the best results on BLEU-2, BLEU-3, and BLEU-4, showing its strong advantages in generating continuous medical phrases, disease-related descriptive fragments, and complex sentence structures. Although it does not achieve the highest scores on some metrics such as BLEU-1, ROUGE-L, METEOR, and CIDEr, its overall performance remains highly competitive. In particular, the advantages on high-order BLEU metrics demonstrate that the proposed method can generate report contents that are more consistent with medical writing conventions and semantically more coherent.

Overall, MAC-RRG achieves competitive NLG performance on the IU X-Ray, CheXpert Plus, and MIMIC-CXR datasets. On the relatively small-scale IU X-Ray dataset, the proposed method shows prominent performance in high-order semantic matching and report content consistency. On the more complex CheXpert Plus and MIMIC-CXR datasets, the proposed method still maintains stable generation quality and achieves the best or near-best results on multiple key metrics. These experimental results fully verify the effectiveness and robustness of the proposed method across datasets with different scales and clinical complexities. They also demonstrate that the knowledge-enhanced strategy based on graph path retrieval and knowledge base retrieval can help the model generate more accurate, coherent, and medically reasonable automated radiology reports.

\noindent $\bullet$ \textbf{Analysis of CE Metric.~} As shown in Table~\ref{tab:ce_metrics}, we further evaluate the clinical correctness of different medical report generation models on the CheXpert Plus dataset using Clinical Efficacy (CE) metrics, including F1 score, Precision, and Recall. These metrics measure whether the generated reports can accurately capture clinically relevant abnormalities, which is crucial for medical report generation.
Compared with existing methods, the proposed MAC-RRG achieves the best results in terms of F1 score and Precision, reaching 0.295 and 0.350, respectively. Specifically, MAC-RRG improves the F1 score from 0.260 achieved by R2GenGPT to 0.295. Meanwhile, its Precision also surpasses that of VLCI, which obtains the highest Precision among the baseline methods. This indicates that our method is more effective in generating clinically accurate pathological findings and reducing false-positive descriptions in the generated reports. Although ORGan achieves the highest Recall of 0.287, its Precision and F1 score are only 0.288 and 0.277, respectively, which are clearly lower than those of MAC-RRG. In contrast, our method achieves a competitive Recall of 0.274 while maintaining the highest Precision and overall F1 score. This demonstrates that MAC-RRG achieves a better balance between correctly identifying pathological findings and avoiding inaccurate clinical statements. Overall, the CE metric results demonstrate that the proposed method can effectively identify key clinical abnormalities from medical images and generate reports with stronger clinical consistency. The higher F1 score further verifies the effectiveness of MAC-RRG in balancing sensitivity and precision, indicating its potential for clinically reliable automatic medical report generation.

\begin{figure}
    \centering
    \includegraphics[width=\linewidth]{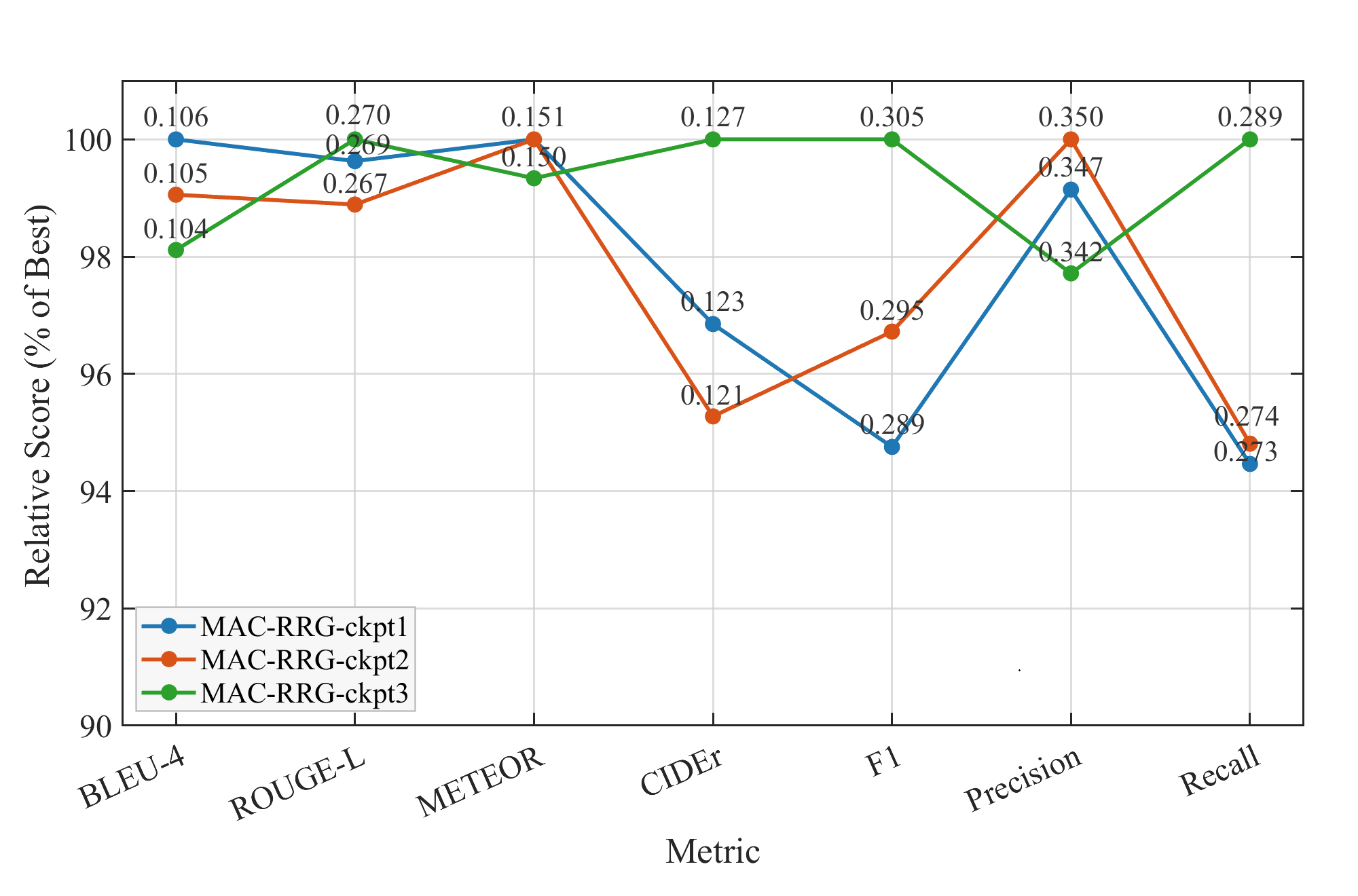}
    \caption{Comparison of different checkpoints on BLEU-4, ROUGE-L, METEOR, CIDEr, F1, Precision, and Recall.}
    \label{fig:report_ablation_checkpoint}
\end{figure}

\subsection{Ablation Study}

\noindent $\bullet$ \textbf{Component Analysis.~} To further validate the effectiveness of each key component in the proposed MAC-RRG, we conduct component analysis experiments on the CheXpert Plus dataset. As shown in Table~\ref{tab:ablation_nlg_chexpertplus}, we mainly evaluate the contributions of three core components, including the MM-KG Agent, the Knowledge Agent, and the Knowledge Fusion Module. Specifically, the MM-KG Agent introduces structured radiological relations by expanding extracted medical entities into local knowledge graph neighborhoods, while the Knowledge Agent retrieves entity-level textual evidence from an external medical knowledge base. The Knowledge Fusion Module further integrates these two complementary knowledge sources into compact representations, reducing the redundancy of long textual prompts and improving knowledge utilization during report generation.

In terms of NLG metrics, both the MM-KG Agent and the Knowledge Agent consistently improve over the baseline. Specifically, the baseline achieves BLEU-1, BLEU-2, BLEU-3, BLEU-4, ROUGE-L, METEOR, and CIDEr scores of 0.361, 0.224, 0.149, 0.101, 0.266, 0.145, and 0.123, respectively. With only the MM-KG Agent, these scores improve to 0.366, 0.227, 0.150, 0.103, 0.266, 0.148, and 0.124, while using only the Knowledge Agent yields 0.370, 0.228, 0.151, 0.103, 0.266, 0.149, and 0.125. When all three components are jointly used, the model achieves the best overall performance on most NLG metrics, with BLEU-1, BLEU-2, BLEU-3, BLEU-4, ROUGE-L, and METEOR scores further improved to 0.375, 0.232, 0.154, 0.105, 0.267, and 0.151, respectively. This suggests that the collaboration of structured knowledge encoding, entity-level knowledge retrieval, and the knowledge fusion module can effectively improve the linguistic quality and semantic consistency of generated reports. Similar improvements are also observed in clinical efficacy metrics, where F1, Precision, and Recall increase from 0.260, 0.315, and 0.224 in the baseline to 0.295, 0.350, and 0.274 in the full model. These results demonstrate that the MM-KG Agent and the Knowledge Agent provide complementary structured and textual medical knowledge, while the Knowledge Fusion Module effectively integrates them, thereby improving the linguistic quality, semantic consistency, and clinical reliability of generated reports.

\noindent $\bullet$ \textbf{Analysis of Checkpoint Selection Strategy.~} As shown in Fig.~\ref{fig:report_ablation_checkpoint}, the final MAC-RRG model is selected based on a balanced consideration of BLEU-4 and clinically relevant Precision, Recall, and F1, rather than the best CIDEr score alone. Specifically, MAC-RRG-ckpt1 achieves the highest BLEU-4 score of 0.106, but its F1 and Recall are relatively lower at 0.289 and 0.273. MAC-RRG-ckpt3 obtains the best CIDEr, ROUGE-L, F1, and Recall scores of 0.127, 0.270, 0.305, and 0.289, respectively, but its BLEU-4 and Precision decrease to 0.104 and 0.342, indicating a tendency toward broader yet less precise clinical descriptions. By comparison, MAC-RRG-ckpt2 achieves a competitive BLEU-4 score of 0.105 and the highest Precision of 0.350, while also outperforming ckpt1 in F1 and Recall. Therefore, we adopt MAC-RRG-ckpt2 as the final model, since it provides a better trade-off between language quality and clinical factual accuracy.

\noindent $\bullet$ \textbf{Analysis on different numbers of neighbors.~} To analyze the influence of different neighborhood scales in the knowledge graph, we conduct ablation experiments by varying the number of retrieved neighboring nodes. As shown in Table~\ref{tab:neighbors_ablation}, when $\text{top}_K$ ranges from 3 to 20, the overall NLG performance remains relatively stable, with BLEU-4, ROUGE-L, METEOR, and CIDEr varying within 0.102--0.105, 0.266--0.269, 0.146--0.151, and 0.121--0.131, respectively. Increasing $\text{top}_K$ from 3 to 10 improves BLEU-4 from 0.102 to 0.105 and METEOR from 0.147 to 0.151, indicating that a moderate number of neighbors can provide useful structured relational information. When $\text{top}_K$ is further increased to 15, BLEU-4 and METEOR decrease to 0.102 and 0.146, respectively. At $\text{top}_K{=}20$, BLEU-4 and METEOR slightly recover to 0.103 and 0.148, while ROUGE-L and CIDEr reach relatively high values of 0.269 and 0.131. These results suggest that although more neighbors may increase knowledge coverage, they may also introduce redundant or weakly relevant information. Overall, $\text{top}_K{=}10$ achieves the best BLEU-4 and METEOR scores while maintaining competitive ROUGE-L performance, and is therefore adopted in the final model.

\begin{table}[tb!]
\centering
\small 
\caption{Compare the effects of different numbers of neighbors.}
\label{tab:neighbors_ablation}
\begin{tabular}{l c c c c}
\hline \toprule [0.5 pt] 
\textbf{\#top\_K} & \textbf{BLEU-4} & \textbf{ROUGE-L} & \textbf{METEOR} & \textbf{CIDEr} \\
\hline
3   & 0.102 & 0.269 &  0.147 & 0.131 \\
5  & 0.103 & 0.267 & 0.148 &  0.128 \\
10  & 0.105 & 0.267 & 0.151 & 0.121 \\
15  & 0.102 & 0.266 &  0.146 &  0.129 \\
20  & 0.103 & 0.269 &  0.148 &  0.131 \\
\hline \toprule [0.5 pt] 
\end{tabular}
\end{table}

\noindent $\bullet$ \textbf{Analysis of contextual entity types.} 
As shown in Table~\ref{tab:entity_types_ablation}, different entity types have different effects on report generation performance. When only anatomy entities are used, the model achieves the highest CIDEr score of 0.133, indicating that anatomy entities provide stable spatial localization information. After introducing disorder entities, the A + D setting obtains the best BLEU-4 and METEOR scores, reaching 0.105 and 0.151, respectively, improving over the anatomy-only setting with 0.102 and 0.148. This suggests that disease-related entities help enhance the description of abnormal findings. In contrast, further adding device entities reduces CIDEr from 0.121 to 0.115. When all entity types are used, ROUGE-L and METEOR also decrease to 0.265 and 0.147, respectively, indicating that excessive entity types may introduce noise or weakly relevant information. Therefore, we adopt Anatomy + Disorder as the default entity-type setting to achieve a better balance between anatomical localization and lesion-level semantic description. 

\begin{table}[tb!]
\centering
\small 
\caption{Compare the effects of contextual entity types.}
\label{tab:entity_types_ablation}
\begin{tabular}{l c c c c}
\hline \toprule [0.5 pt] 
\textbf{\#entity\_types} & \textbf{BLEU-4} & \textbf{ROUGE-L} & \textbf{METEOR} & \textbf{CIDEr} \\
\hline
A   & 0.102 & 0.269 &  0.148 &  0.133 \\
A + D  & 0.105 & 0.267 & 0.151 & 0.121 \\
A + D+ Dev  & 0.103 & 0.266 &  0.149 &  0.115 \\
All & 0.103 & 0.265 &  0.147 &  0.120 \\
\hline \toprule [0.5 pt] 
\end{tabular}
\end{table}

\noindent $\bullet$ \textbf{Analysis of different numbers of chunks.}
We investigate the effect of retrieval corpus size by varying the nominal number of knowledge chunks from 200 to 1,500 while keeping all other settings unchanged. After preprocessing and deduplication, the nominal 400-chunk configuration contains 393 valid chunks. As shown in Table~\ref{tab:chunks_ablation}, increasing the corpus size from 200 to 400 generally improves generation performance, with the 400-chunk setting achieving the best or comparable ROUGE-L and METEOR scores and competitive BLEU-4 and CIDEr results. Checkpoint selection jointly considers NLG and clinical efficacy metrics rather than BLEU-4 alone. Although another checkpoint under the 400-chunk setting also reaches a BLEU-4 score of 0.106, it is not selected because of its lower Precision, Recall, and F1 scores. Similarly, the 1,000-chunk setting achieves a BLEU-4 score of 0.106 but obtains only 0.278, 0.334, and 0.263 in F1, Precision, and Recall, respectively, lower than the selected 400-chunk checkpoint (0.295, 0.350, and 0.274). Moreover, increasing the corpus size to 1,500 chunks degrades all NLG metrics, possibly due to redundant or weakly relevant evidence. Therefore, we adopt the nominal 400-chunk configuration to balance linguistic quality and clinical efficacy.

\begin{table}[tb!]
\centering
\small 
\caption{Compare the effects of different numbers of chunks.}
\label{tab:chunks_ablation}
\begin{tabular}{l c c c c}
\hline \toprule [0.5 pt] 
\textbf{\#chunks} & \textbf{BLEU-4} & \textbf{ROUGE-L} & \textbf{METEOR} & \textbf{CIDEr} \\
\hline
200   & 0.104 & 0.266 & 0.150 & 0.121 \\
300  & 0.104 & 0.266 & 0.148 & 0.122 \\
400  & 0.105 & 0.267 & 0.151 & 0.121 \\
1000  & 0.106 & 0.266 & 0.150 & 0.123 \\
1500  & 0.103 & 0.263 & 0.146 & 0.121 \\
\hline \toprule [0.5 pt] 
\end{tabular}
\end{table}

\noindent $\bullet$ \textbf{Analysis of different fusion strategies.~}  
The final version of this study adopts direct concatenation as the knowledge fusion strategy, rather than introducing additional fusion through self-attention or cross-attention, mainly based on the experimental results. As shown in Table~\ref{tab:fusion_strategy}, the concat strategy achieves the best performance on BLEU-4 and METEOR, while maintaining competitive results on ROUGE-L. In contrast, self-attention and cross-attention do not bring stable improvements.

This indicates that the graph branch and the retrieval branch have already formed effective continuous knowledge representations after graph attention encoding and Bio-ClinicalBERT encoding. Introducing additional complex attention interactions may increase the number of parameters and amplify the influence of retrieval noise or weakly relevant neighboring nodes in the knowledge graph. Therefore, direct concatenation can preserve graph-structured knowledge, retrieved textual knowledge, and visual information more stably while maintaining a simple architecture, thereby enabling effective enhancement of the large language model.

\begin{table}[tb!]
\centering
\small 
\caption{Compare the effects of different fusion strategies}
\label{tab:fusion_strategy}
\begin{tabular}{l c c c c}
\hline \toprule [0.5 pt] 
\textbf{\#strategies} & \textbf{BLEU-4} & \textbf{ROUGE-L} & \textbf{METEOR} & \textbf{CIDEr} \\
\hline
self\_attention   & 0.102 & 0.266 &  0.148 &  0.125 \\
concat  & 0.105 & 0.267 & 0.151 & 0.121 \\
cross\_attention  & 0.104 & 0.267 & 0.149 &  0.116 \\
\hline \toprule [0.5 pt] 
\end{tabular}
\end{table}

\begin{figure*}[!htp]
\newcommand{\MatchingKGGpt}[1]{\sethlcolor{pink}\hl{#1}}
\newcommand{\MatchingGpt}[1]{\sethlcolor{cyan}\hl{#1}}
\newcommand{\MatchingKG}[1]{\sethlcolor{yellow}\hl{#1}}
\centering
\resizebox{\textwidth}{!}{%
\setlength{\tabcolsep}{3pt}%
\begin{tabular}{@{} >{\centering\arraybackslash}p{4cm} p{7cm} p{7cm} p{7cm} @{}}
\toprule
\textbf{Image} & \multicolumn{1}{c}{\textbf{Ground Truth}} & \multicolumn{1}{c}{\textbf{Ours}} & \multicolumn{1}{c}{\textbf{R2GenGPT}} \\
\midrule
\vspace{0pt}\includegraphics[width=3.5cm]{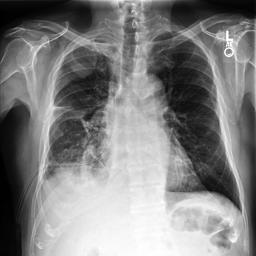}
& \vspace{0pt}\MatchingKG{Loculated pleural effusions are again seen.} \MatchingKG{A rounded pleural-based opacity posteriorly laterally at the right lung apex also likely reflects a loculated effusion.} \MatchingKG{Patchy opacities are also seen throughout the right lung zone, most prominent the right lung base, which may reflect adjacent atelectasis or consolidation.} \MatchingKG{The left lung appears largely clear.} \MatchingGpt{The cardiomediastinal silhouette appears unchanged.} \MatchingKGGpt{No acute bony abnormalities are identified.} \MatchingKG{Multilevel degenerative changes are again seen of the thoracic spine.}
& \vspace{0pt}postsurgical changes of the right hemithorax are again seen. \MatchingKG{there is a small right pleural effusion with associated atelectasis or consolidation at the right base.} \MatchingKG{the left lung is clear.} the heart is normal in size. the mediastinal contours are normal. \MatchingKGGpt{no acute osseous abnormalities are identified.} \MatchingKG{degenerative changes of the thoracic spine are again seen.}
& \vspace{0pt}\MatchingGpt{the cardiomediastinal silhouette is unchanged.} there is redemonstration of a small right apical pneumothorax which has decreased in size since the prior examination. there is persistent elevation of the right hemidiaphragm with associated blunting of the right costophrenic angle. there is no evidence of pneumothorax or pleural effusion on the left side. \MatchingKGGpt{there are no osseous abnormalities.} surgical clips are again seen in the right upper quadrant. \\[8pt]

\vspace{0pt}\includegraphics[width=3.5cm]{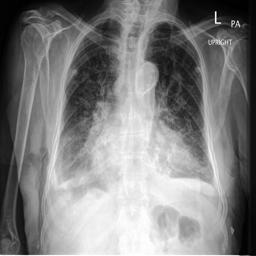}
& \vspace{0pt}The cardiac silhouette is enlarged. \MatchingKG{Atherosclerotic calcifications of the aorta.} \MatchingKG{Prominent vascular markings consistent with pulmonary edema.} \MatchingKGGpt{Blunting of the bilateral costophrenic angles, likely small pleural effusions.} \MatchingKG{Crowding of the vascular markings in the bilateral bases.} \MatchingKG{Osteopenia.}
& \vspace{0pt}\MatchingKG{single ap view of the chest demonstrates diffuse increased interstitial markings consistent with pulmonary edema.} \MatchingKGGpt{there are small bilateral pleural effusions.} \MatchingKGGpt{in addition there is blunting of the costophrenic angles consistent with small bilateral pleural effusions.} the cardiomediastinal silhouette is unremarkable. \MatchingKG{there is atherosclerotic calcification of the aortic arch and thoracic aorta.} \MatchingKG{bones demonstrate osteopenia.} no soft tissue abnormalities are identified.
& \vspace{0pt}the cardiomediastinal silhouette is within normal limits. there are bibasilar opacities which may represent atelectasis or consolidation. \MatchingKGGpt{there are also bilateral pleural effusions.} the visualized osseous structures are unremarkable. no acute bony abnormalities are identified. degenerative changes are seen in the thoracic spine. \\[8pt]

\vspace{0pt}\includegraphics[width=3.5cm]{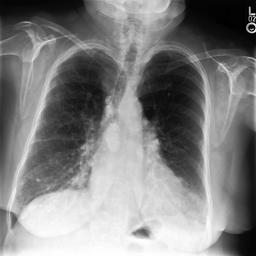}
& \vspace{0pt}\MatchingKGGpt{Cardiac silhouette is not enlarged.} \MatchingKGGpt{Mediastinal contour is normal.} Multiple calcified mediastinal and hilar nodes. \MatchingKG{Abnormal hazy and nodular opacities at both lung bases are concerning for pneumonia or aspiration.} Possible emphysema. \MatchingKG{No acute bony abnormalities.} \MatchingKG{No pneumothorax.}
& \vspace{0pt}\MatchingKGGpt{the cardiomediastinal silhouette is normal.} \MatchingKG{increased interstitial opacities are noted at the bilateral lung bases.} no pleural effusions. \MatchingKG{no pneumothorax.} multilevel degenerative changes of the thoracic spine. surgical clips are seen in the right axilla. \MatchingKG{there are no acute osseous abnormalities.} biapical pleural thickening is noted on the lateral radiograph.
& \vspace{0pt}pa and lateral chest radiographs are obtained. \MatchingKGGpt{the cardiomediastinal silhouette is within normal limits.} there is mild prominence of the pulmonary vasculature. there is no focal consolidation or pleural effusion. there is blunting of the costophrenic angles bilaterally which may represent small pleural effusions. degenerative changes are seen in the thoracic spine. \\
\bottomrule
\end{tabular}
}%
\caption{X-ray images and their corresponding ground-truth reports, along with the generated reports of MAC-RRG and R2GenGPT on the CheXpert Plus dataset. Matching sentences in our report are highlighted in yellow, R2GenGPT matching sentences are highlighted in cyan, and sentences matched by both models are highlighted in pink.}
\label{fig:visual_report}
\end{figure*}

\subsection{Visualization} 
\noindent $\bullet$ \textbf{Visualization of Structured and Textual Knowledge Evidence.~} 
As shown in Fig.~\ref{fig:kg_chunks}, we select the chest X-ray image
\texttt{train\_patient13647\_study2\_view1\_frontal} and its corresponding
Draft Report as a representative case to illustrate how the proposed
collaborative agents retrieve and organize complementary knowledge. The
upper part of the figure presents the input chest radiograph together with
the initial Draft Report, which describes cardiomediastinal enlargement,
mild pulmonary vascular prominence with possible pulmonary edema, a small
left pleural effusion, left costophrenic angle blunting with possible
atelectasis, absence of pneumothorax, thoracic aortic calcification, and
degenerative changes of the thoracic spine.

Given the Draft Report, the MM-KG Agent first identifies case-relevant
contextual entities and retrieves their associated triples from the medical
knowledge graph. For clarity and interpretability, only a subset of
representative triples is visualized in the lower-left panel. These triples
are organized into five independent entity-centered subgraphs corresponding
to cardiomediastinal findings, atelectasis, pleural effusion, thoracic
anatomical structures, and pulmonary edema. The relations
\texttt{LOCATED\_AT}, \texttt{MODIFY}, and
\texttt{SUGGESTIVE\_OF} encode anatomical associations, modifier
relationships, and diagnostic associations between radiographic findings
and potential abnormalities, respectively. Consequently, medical concepts
that are dispersed throughout the Draft Report are reorganized into
structured and traceable graph-based evidence.

In parallel, the Knowledge Agent retrieves semantically relevant text chunks
from an external knowledge base, as shown in the lower-right panel. The
representative chunks K1--K3 contain radiological descriptions related to
cardiomegaly, pulmonary vascular congestion or pulmonary edema, pleural
effusion, costophrenic angle blunting, and lower-lung abnormalities. In
contrast to the concise relational evidence provided by the MM-KG Agent,
these chunks supply richer descriptive context and broader co-occurrence
patterns for the findings mentioned in the Draft Report.

Overall, the visualization demonstrates the complementary roles of the two
knowledge agents. The MM-KG Agent provides structured evidence at the entity
and relation levels, whereas the Knowledge Agent contributes descriptive
textual evidence. These complementary knowledge sources jointly support the
subsequent knowledge-fusion module and facilitate the refinement of the
initial radiology report.

\noindent $\bullet$ \textbf{Report Generation.~} 
As shown in Figure~\ref{fig:visual_report}, we present qualitative examples from the CheXpert Plus dataset to further demonstrate the effectiveness of the proposed MAC-RRG model for X-ray image-based radiology report generation. For each X-ray image, we compare the ground-truth report with the reports generated by MAC-RRG and the baseline model R2GenGPT. To make the comparison more intuitive, sentences in the reports generated by our model that match the ground truth are highlighted in yellow, those generated by R2GenGPT that match the ground truth are highlighted in cyan, and sentences correctly matched by both models are highlighted in pink.

The visualization results show that MAC-RRG achieves stronger consistency with the ground-truth reports in both positive findings and negative observations. This indicates that, compared with R2GenGPT, the reports generated by MAC-RRG are more semantically aligned with the ground truth. These qualitative results further demonstrate that the proposed multi-agent collaborative framework can improve the clinical completeness and factual consistency of generated radiology reports.

\subsection{Limitation Analysis}
Although MAC-RRG improves radiology report generation by incorporating structured knowledge graph evidence and entity-level retrieved textual knowledge, its performance is still influenced by the quality of the initial draft report. Since contextual entities are extracted from the preliminary report, missed or incorrectly described abnormalities may lead the MM-KG Agent and Knowledge Agent to retrieve incomplete or less relevant external evidence. This dependency indicates that more robust entity extraction and draft correction mechanisms remain important for further improving iterative report refinement.

The current framework adopts direct concatenation as the final knowledge fusion strategy after projecting graph and retrieval representations into the visual feature space. We also investigated more complex fusion strategies, including self-attention and cross-attention. However, as reported in Table~\ref{tab:fusion_strategy}, these strategies do not consistently improve the overall performance under the current setting. Self-attention achieves a slightly higher CIDEr score but reduces BLEU-4 and METEOR, while cross-attention obtains comparable ROUGE-L but leads to a lower CIDEr score. One possible reason is that additional attention interactions may amplify weakly relevant retrieved chunks or noisy neighboring nodes in the knowledge graph, especially when external knowledge has already been encoded into compact continuous representations. Therefore, direct concatenation is adopted in this work as a more stable and lightweight fusion strategy.


\section{Conclusion} \label{sec::conclusion} 
In this paper, we propose MAC-RRG, an iterative multi-agent collaboration framework for X-ray radiology report generation. The core idea is to use the initial draft report as an entity-aware bridge between visual evidence and external medical knowledge. Specifically, the MM-KG agent retrieves structured entity-centered relations from a medical knowledge graph, while the Knowledge agent retrieves complementary entity-level textual evidence from an external knowledge corpus. These two knowledge sources are fused with visual tokens and projected into the LLM embedding space to guide the generation of more clinically consistent reports. Extensive experiments on IU X-Ray, CheXpert Plus, and MIMIC-CXR demonstrate the effectiveness of the proposed framework. MAC-RRG achieves competitive or superior performance on multiple NLG metrics, especially on high-order BLEU scores, indicating its ability to generate more coherent medical phrases and disease-related descriptions. 

Overall, MAC-RRG shows that role-specialized knowledge agents can provide complementary structured and textual evidence for LLM-based radiology report generation. In future work, we will further investigate more robust entity extraction, larger-scale medical knowledge bases, uncertainty-aware retrieval, and adaptive knowledge fusion strategies to improve the generalization and clinical trustworthiness of automatic radiology report generation systems.

\section*{Acknowledgment} \label{sec::acknowledgement} 
This work was supported by the National Natural Science Foundation of China under Grant 62572004, 62102205, U24A20342. Anhui Provincial Natural Science Foundation-Outstanding Youth Project, 2408085Y032. The authors acknowledge the High-performance Computing Platform of Anhui University for providing computing resources.

\small{ 
\bibliographystyle{IEEEtran}
\bibliography{reference}

\begin{thebibliography}{10}
\providecommand{\url}[1]{#1}
\csname url@samestyle\endcsname
\providecommand{\newblock}{\relax}
\providecommand{\bibinfo}[2]{#2}
\providecommand{\BIBentrySTDinterwordspacing}{\spaceskip=0pt\relax}
\providecommand{\BIBentryALTinterwordstretchfactor}{4}
\providecommand{\BIBentryALTinterwordspacing}{\spaceskip=\fontdimen2\font plus
\BIBentryALTinterwordstretchfactor\fontdimen3\font minus
  \fontdimen4\font\relax}
\providecommand{\BIBforeignlanguage}[2]{{%
\expandafter\ifx\csname l@#1\endcsname\relax
\typeout{** WARNING: IEEEtran.bst: No hyphenation pattern has been}%
\typeout{** loaded for the language `#1'. Using the pattern for}%
\typeout{** the default language instead.}%
\else
\language=\csname l@#1\endcsname
\fi
#2}}
\providecommand{\BIBdecl}{\relax}
\BIBdecl

\bibitem{messina2022survey}
P.~Messina, P.~Pino, D.~Parra, A.~Soto, C.~Besa, S.~Uribe, M.~And{\'\i}a,
  C.~Tejos, C.~Prieto, and D.~Capurro, ``A survey on deep learning and
  explainability for automatic report generation from medical images,''
  \emph{ACM Computing Surveys (CSUR)}, vol.~54, no. 10s, pp. 1--40, 2022.

\bibitem{ganeshan2018structured}
D.~Ganeshan, P.-A.~T. Duong, L.~Probyn, L.~Lenchik, T.~A. McArthur,
  M.~Retrouvey, E.~H. Ghobadi, S.~L. Desouches, D.~Pastel, and I.~R. Francis,
  ``Structured reporting in radiology,'' \emph{Academic radiology}, vol.~25,
  no.~1, pp. 66--73, 2018.

\bibitem{lecun1998gradient}
Y.~LeCun, L.~Bottou, Y.~Bengio, and P.~Haffner, ``Gradient-based learning
  applied to document recognition,'' \emph{Proceedings of the IEEE}, vol.~86,
  no.~11, pp. 2278--2324, 1998.

\bibitem{jing2018automatic}
B.~Jing, P.~Xie, and E.~Xing, ``On the automatic generation of medical imaging
  reports,'' in \emph{Proceedings of the 56th annual meeting of the association
  for computational linguistics (volume 1: long papers)}, 2018, pp. 2577--2586.

\bibitem{wang2018tienet}
X.~Wang, Y.~Peng, L.~Lu, Z.~Lu, and R.~M. Summers, ``Tienet: Text-image
  embedding network for common thorax disease classification and reporting in
  chest x-rays,'' in \emph{Proceedings of the IEEE conference on computer
  vision and pattern recognition}, 2018, pp. 9049--9058.

\bibitem{li2018hybrid}
Y.~Li, X.~Liang, Z.~Hu, and E.~P. Xing, ``Hybrid retrieval-generation
  reinforced agent for medical image report generation,'' \emph{Advances in
  neural information processing systems}, vol.~31, 2018.

\bibitem{hochreiter1997long}
S.~Hochreiter and J.~Schmidhuber, ``Long short-term memory,'' \emph{Neural
  computation}, vol.~9, no.~8, pp. 1735--1780, 1997.

\bibitem{ashish2017attention}
V.~Ashish, ``Attention is all you need,'' \emph{Advances in neural information
  processing systems}, 2017.

\bibitem{chen2020generating}
Z.~Chen, Y.~Song, T.-H. Chang, and X.~Wan, ``Generating radiology reports via
  memory-driven transformer,'' in \emph{Proceedings of the 2020 conference on
  empirical methods in natural language processing (EMNLP)}, 2020, pp.
  1439--1449.

\bibitem{chen2021cross}
Z.~Chen, Y.~Shen, Y.~Song, and X.~Wan, ``Cross-modal memory networks for
  radiology report generation,'' in \emph{Proceedings of the 59th annual
  meeting of the association for computational linguistics and the 11th
  international joint conference on natural language processing (volume 1: long
  papers)}, 2021, pp. 5904--5914.

\bibitem{liu2021exploring}
F.~Liu, X.~Wu, S.~Ge, W.~Fan, and Y.~Zou, ``Exploring and distilling posterior
  and prior knowledge for radiology report generation,'' in \emph{Proceedings
  of the IEEE/CVF conference on computer vision and pattern recognition}, 2021,
  pp. 13\,753--13\,762.

\bibitem{wang2023metransformer}
Z.~Wang, L.~Liu, L.~Wang, and L.~Zhou, ``Metransformer: Radiology report
  generation by transformer with multiple learnable expert tokens,'' in
  \emph{Proceedings of the IEEE/CVF conference on computer vision and pattern
  recognition}, 2023, pp. 11\,558--11\,567.

\bibitem{li2023blip}
J.~Li, D.~Li, S.~Savarese, and S.~Hoi, ``Blip-2: Bootstrapping language-image
  pre-training with frozen image encoders and large language models,'' in
  \emph{International conference on machine learning}.\hskip 1em plus 0.5em
  minus 0.4em\relax PmLR, 2023, pp. 19\,730--19\,742.

\bibitem{wang2023r2gengpt}
Z.~Wang, L.~Liu, L.~Wang, and L.~Zhou, ``R2gengpt: Radiology report generation
  with frozen llms,'' \emph{Meta-Radiology}, vol.~1, no.~3, p. 100033, 2023.

\bibitem{wang2026r2gencsr}
X.~Wang, Y.~Li, F.~Wang, S.~Wang, C.~Li, and B.~Jiang, ``R2gencsr: Mining
  contextual and residual information for llms-based radiology report
  generation,'' \emph{IEEE Journal of Biomedical and Health Informatics}, 2026.

\bibitem{wang2025cxpmrg}
X.~Wang, F.~Wang, Y.~Li, Q.~Ma, S.~Wang, B.~Jiang, and J.~Tang, ``Cxpmrg-bench:
  Pre-training and benchmarking for x-ray medical report generation on chexpert
  plus dataset,'' in \emph{2025 IEEE/CVF Conference on Computer Vision and
  Pattern Recognition (CVPR)}.\hskip 1em plus 0.5em minus 0.4em\relax IEEE,
  2025, pp. 5123--5133.

\bibitem{hong2024metagpt}
S.~Hong, M.~Zhuge, J.~Chen, X.~Zheng, Y.~Cheng, J.~Wang, C.~Zhang, S.~Yau,
  Z.~Lin, L.~Zhou \emph{et~al.}, ``Metagpt: Meta programming for a multi-agent
  collaborative framework,'' in \emph{International Conference on Learning
  Representations}, vol. 2024, 2024, pp. 23\,247--23\,275.

\bibitem{wu2023autogen}
Q.~Wu, G.~Bansal, J.~Zhang, Y.~Wu, B.~Li, E.~Zhu, L.~Jiang, X.~Zhang, S.~Zhang,
  J.~Liu \emph{et~al.}, ``Autogen: Enabling next-gen llm applications via
  multi-agent conversation,'' \emph{arXiv preprint arXiv:2308.08155}, 2023.

\bibitem{chen2024towards}
J.~Chen, C.~Gui, R.~Ouyang, A.~Gao, S.~Chen, G.~H. Chen, X.~Wang, Z.~Cai,
  K.~Ji, X.~Wan \emph{et~al.}, ``Towards injecting medical visual knowledge
  into multimodal llms at scale,'' in \emph{Proceedings of the 2024 conference
  on empirical methods in natural language processing}, 2024, pp. 7346--7370.

\bibitem{pellegrini2025radialog}
C.~Pellegrini, E.~{\"O}zsoy, B.~Busam, B.~Wiestler, N.~Navab, and M.~Keicher,
  ``Radialog: Large vision-language models for x-ray reporting and
  dialog-driven assistance,'' in \emph{Medical imaging with deep learning},
  2025.

\bibitem{li2019knowledge}
C.~Y. Li, X.~Liang, Z.~Hu, and E.~P. Xing, ``Knowledge-driven encode, retrieve,
  paraphrase for medical image report generation,'' in \emph{Proceedings of the
  AAAI conference on artificial intelligence}, vol.~33, no.~01, 2019, pp.
  6666--6673.

\bibitem{huang2023kiut}
Z.~Huang, X.~Zhang, and S.~Zhang, ``Kiut: Knowledge-injected u-transformer for
  radiology report generation,'' in \emph{Proceedings of the IEEE/CVF
  conference on computer vision and pattern recognition}, 2023, pp.
  19\,809--19\,818.

\bibitem{li2024kargen}
Y.~Li, Z.~Wang, Y.~Liu, L.~Wang, L.~Liu, and L.~Zhou, ``Kargen:
  Knowledge-enhanced automated radiology report generation using large language
  models,'' in \emph{International Conference on Medical Image Computing and
  Computer-Assisted Intervention}.\hskip 1em plus 0.5em minus 0.4em\relax
  Springer, 2024, pp. 382--392.

\bibitem{jain2021radgraph}
S.~Jain, A.~Agrawal, A.~Saporta, S.~Truong, D.~N. Duong, T.~Bui, P.~Chambon,
  Y.~Zhang, M.~P. Lungren, A.~Y. Ng \emph{et~al.}, ``Radgraph: Extracting
  clinical entities and relations from radiology reports (2021),'' \emph{arXiv
  preprint arXiv:2106.14463}, 2021.

\bibitem{wu2021chest}
J.~T. Wu, N.~N. Agu, I.~Lourentzou, A.~Sharma, J.~A. Paguio, J.~S. Yao, E.~C.
  Dee, W.~G. Mitchell, S.~Kashyap, A.~Giovannini \emph{et~al.}, ``Chest
  imagenome dataset for clinical reasoning,'' in \emph{Thirty-fifth Conference
  on Neural Information Processing Systems Datasets and Benchmarks Track (Round
  2)}, 2021.

\bibitem{lewis2020retrieval}
P.~Lewis, E.~Perez, A.~Piktus, F.~Petroni, V.~Karpukhin, N.~Goyal,
  H.~K{\"u}ttler, M.~Lewis, W.-t. Yih, T.~Rockt{\"a}schel \emph{et~al.},
  ``Retrieval-augmented generation for knowledge-intensive nlp tasks,''
  \emph{Advances in neural information processing systems}, vol.~33, pp.
  9459--9474, 2020.

\bibitem{ranjit2023retrieval}
M.~Ranjit, G.~Ganapathy, R.~Manuel, and T.~Ganu, ``Retrieval augmented chest
  x-ray report generation using openai gpt models,'' in \emph{Machine learning
  for healthcare conference}.\hskip 1em plus 0.5em minus 0.4em\relax PMLR,
  2023, pp. 650--666.

\bibitem{song2024lab}
S.~Song, A.~Subramanyam, I.~Madejski, and R.~L. Grossman, ``Lab-rag: Label
  boosted retrieval augmented generation for radiology report generation,''
  \emph{arXiv preprint arXiv:2411.16523}, 2024.

\bibitem{park2026ra}
J.~Park, B.~Yoon, S.~Kim, and K.~Choi, ``Ra-rrg: Multimodal retrieval-augmented
  radiology report generation with key phrase extraction,'' in \emph{Findings
  of the Association for Computational Linguistics: ACL 2026}, 2026, pp.
  5029--5048.

\bibitem{yao2022react}
S.~Yao, J.~Zhao, D.~Yu, N.~Du, I.~Shafran, K.~Narasimhan, and Y.~Cao, ``React:
  Synergizing reasoning and acting in language models,'' \emph{arXiv preprint
  arXiv:2210.03629}, 2022.

\bibitem{zhang2020radiology}
Y.~Zhang, X.~Wang, Z.~Xu, Q.~Yu, A.~Yuille, and D.~Xu, ``When radiology report
  generation meets knowledge graph,'' in \emph{Proceedings of the AAAI
  conference on artificial intelligence}, vol.~34, no.~07, 2020, pp.
  12\,910--12\,917.

\bibitem{you2021aligntransformer}
D.~You, F.~Liu, S.~Ge, X.~Xie, J.~Zhang, and X.~Wu, ``Aligntransformer:
  Hierarchical alignment of visual regions and disease tags for medical report
  generation,'' in \emph{International Conference on Medical Image Computing
  and Computer-Assisted Intervention}.\hskip 1em plus 0.5em minus 0.4em\relax
  Springer, 2021, pp. 72--82.

\bibitem{liu2021competence}
F.~Liu, S.~Ge, and X.~Wu, ``Competence-based multimodal curriculum learning for
  medical report generation,'' in \emph{Proceedings of the 59th Annual Meeting
  of the Association for Computational Linguistics and the 11th International
  Joint Conference on Natural Language Processing (Volume 1: Long Papers)},
  2021, pp. 3001--3012.

\bibitem{li2023DCL}
M.~Li, B.~Lin, Z.~Chen, H.~Lin, X.~Liang, and X.~Chang, ``Dynamic graph
  enhanced contrastive learning for chest x-ray report generation,'' in
  \emph{Proceedings of the IEEE/CVF Conference on Computer Vision and Pattern
  Recognition}, 2023, pp. 3334--3343.

\bibitem{jin2024promptmrg}
H.~Jin, H.~Che, Y.~Lin, and H.~Chen, ``Promptmrg: Diagnosis-driven prompts for
  medical report generation,'' in \emph{Proceedings of the AAAI Conference on
  Artificial Intelligence}, vol.~38, no.~3, 2024, pp. 2607--2615.

\bibitem{liu2024multi}
A.~Liu, Y.~Guo, J.-h. Yong, and F.~Xu, ``Multi-grained radiology report
  generation with sentence-level image-language contrastive learning,''
  \emph{IEEE Transactions on Medical Imaging}, vol.~43, no.~7, pp. 2657--2669,
  2024.

\bibitem{fang2025automated}
J.~Fang, S.~Xing, K.~Li, Z.~Guo, G.~Li, and C.~Yu, ``Automated generation of
  chest x-ray imaging diagnostic reports by multimodal and multi granularity
  features fusion,'' \emph{Biomedical Signal Processing and Control}, vol. 105,
  p. 107562, 2025.

\bibitem{rahman2025duco}
Z.~U. Rahman, J.-H. Lee, D.~T. Vu, I.~Murtza, and J.-Y. Kim, ``Duco-net:
  Dual-contrastive learning network for medical report retrieval leveraging
  enhanced encoders and augmentations,'' \emph{IEEE Access}, 2025.

\bibitem{wu2026disease}
P.~Wu, H.~Dong, Y.~Lin, Y.~Ding, and Y.~Peng, ``A disease-aware dual-stage
  framework for chest x-ray report generation,'' in \emph{Proceedings of the
  AAAI Conference on Artificial Intelligence}, vol.~40, no.~40, 2026, pp.
  33\,953--33\,961.

\bibitem{sun2026dvpalign}
D.~Sun, Z.~Chen, C.~Mu, Y.~Liu, C.~Dong, and B.~Luo, ``Dvpalign: Dual-flow
  visual graph encoding with diagnostic-driven prompt alignment for medical
  report generation,'' in \emph{ICASSP 2026-2026 IEEE International Conference
  on Acoustics, Speech and Signal Processing (ICASSP)}.\hskip 1em plus 0.5em
  minus 0.4em\relax IEEE, 2026, pp. 6581--6585.

\bibitem{wang2022cross}
J.~Wang, A.~Bhalerao, and Y.~He, ``Cross-modal prototype driven network for
  radiology report generation,'' in \emph{European Conference on Computer
  Vision}.\hskip 1em plus 0.5em minus 0.4em\relax Springer, 2022, pp. 563--579.

\bibitem{hou2023organ}
W.~Hou, K.~Xu, Y.~Cheng, W.~Li, and J.~Liu, ``Organ: Observation-guided
  radiology report generation via tree reasoning,'' in \emph{Proceedings of the
  61st annual meeting of the association for computational linguistics (volume
  1: Long papers)}, 2023, pp. 8108--8122.

\bibitem{xue2024generating}
Y.~Xue, Y.~Tan, L.~Tan, J.~Qin, and X.~Xiang, ``Generating radiology reports
  via auxiliary signal guidance and a memory-driven network,'' \emph{Expert
  Systems with Applications}, vol. 237, p. 121260, 2024.

\bibitem{yang2024token}
Y.~Yang, J.~Yu, Z.~Fu, K.~Zhang, T.~Yu, X.~Wang, H.~Jiang, J.~Lv, Q.~Huang, and
  W.~Han, ``Token-mixer: Bind image and text in one embedding space for medical
  image reporting,'' \emph{IEEE Transactions on Medical Imaging}, vol.~43,
  no.~11, pp. 4017--4028, 2024.

\bibitem{xing2025mca}
Q.~Xing, Z.~Song, Y.~Zhang, N.~Feng, J.~Yu, and W.~Yang, ``Mca-rg: Enhancing
  llms with medical concept alignment for radiology report generation,'' in
  \emph{International Conference on Medical Image Computing and
  Computer-Assisted Intervention}.\hskip 1em plus 0.5em minus 0.4em\relax
  Springer, 2025, pp. 380--390.

\bibitem{chen2025cross}
W.~Chen, Y.~Liu, C.~Wang, J.~Zhu, G.~Li, C.-L. Liu, and L.~Lin, ``Cross-modal
  causal representation learning for radiology report generation,'' \emph{IEEE
  Transactions on Image Processing}, 2025.

\bibitem{clinicalBert}
B.~Yan and M.~Pei, ``Clinical-bert: Vision-language pre-training for radiograph
  diagnosis and reports generation,'' in \emph{Proceedings of the AAAI
  Conference on Artificial Intelligence}, vol.~36, no.~3, 2022, pp. 2982--2990.

\bibitem{chen2024AdaMatch_Cyclic}
W.~Chen, L.~Shen, J.~Lin, J.~Luo, X.~Li, and Y.~Yuan, ``Fine-grained image-text
  alignment in medical imaging enables explainable cyclic image-report
  generation,'' in \emph{Proceedings of the 62nd Annual Meeting of the
  Association for Computational Linguistics (Volume 1: Long Papers)}, 2024, pp.
  9494--9509.

\bibitem{lang2025dacg}
W.~Lang, Z.~Liu, and Y.~Zhang, ``Dacg: Dual attention and context guidance
  model for radiology report generation,'' \emph{Medical Image Analysis},
  vol.~99, p. 103377, 2025.

\bibitem{10770258}
J.~Zhao, Y.~Zhou, Z.~Chen, H.~Fu, and L.~Wan, ``Topicwise separable sentence
  retrieval for medical report generation,'' \emph{IEEE Transactions on Medical
  Imaging}, vol.~44, no.~3, pp. 1505--1517, 2025.

\bibitem{tang2026graph}
M.~Tang, C.~Tang, J.~Kong, D.~Wang, and T.~Lu, ``Graph-augmented topological
  internalization with dual-stream classifiers for medical report generation,''
  \emph{arXiv preprint arXiv:2605.02376}, 2026.

\bibitem{demner2016iuxray}
D.~Demner-Fushman, M.~D. Kohli, M.~B. Rosenman, S.~E. Shooshan, L.~Rodriguez,
  S.~Antani, G.~R. Thoma, and C.~J. McDonald, ``Preparing a collection of
  radiology examinations for distribution and retrieval,'' \emph{Journal of the
  American Medical Informatics Association}, vol.~23, no.~2, pp. 304--310,
  2016.

\bibitem{johnson2019mimicCXR}
A.~E. Johnson, T.~J. Pollard, S.~J. Berkowitz, N.~R. Greenbaum, M.~P. Lungren,
  C.-y. Deng, R.~G. Mark, and S.~Horng, ``Mimic-cxr, a de-identified publicly
  available database of chest radiographs with free-text reports,''
  \emph{Scientific data}, vol.~6, no.~1, p. 317, 2019.

\bibitem{chambon2024CheXpertPLUS}
P.~Chambon, J.-B. Delbrouck, T.~Sounack, S.-C. Huang, Z.~Chen, M.~Varma, S.~Q.
  Truong, C.~T. Chuong, and C.~P. Langlotz, ``Chexpert plus: Augmenting a large
  chest x-ray dataset with text radiology reports, patient demographics and
  additional image formats,'' \emph{arXiv preprint arXiv:2405.19538}, 2024.

\bibitem{papineni2002bleu}
K.~Papineni, S.~Roukos, T.~Ward, and W.-J. Zhu, ``Bleu: a method for automatic
  evaluation of machine translation,'' in \emph{Proceedings of the 40th annual
  meeting of the Association for Computational Linguistics}, 2002, pp.
  311--318.

\bibitem{lin2004rouge}
C.-Y. Lin, ``Rouge: A package for automatic evaluation of summaries,'' in
  \emph{Text summarization branches out}, 2004, pp. 74--81.

\bibitem{banerjee2005meteor}
S.~Banerjee and A.~Lavie, ``Meteor: An automatic metric for mt evaluation with
  improved correlation with human judgments,'' in \emph{Proceedings of the acl
  workshop on intrinsic and extrinsic evaluation measures for machine
  translation and/or summarization}, 2005, pp. 65--72.

\bibitem{vedantam2015cider}
R.~Vedantam, C.~Lawrence~Zitnick, and D.~Parikh, ``Cider: Consensus-based image
  description evaluation,'' in \emph{Proceedings of the IEEE conference on
  computer vision and pattern recognition}, 2015, pp. 4566--4575.

\bibitem{yan2021weakly}
A.~Yan, Z.~He, X.~Lu, J.~Du, E.~Chang, A.~Gentili, J.~McAuley, and C.-N. Hsu,
  ``Weakly supervised contrastive learning for chest x-ray report generation,''
  in \emph{Findings of the association for computational linguistics: EMNLP
  2021}, 2021, pp. 4009--4015.

\bibitem{liu2021swin}
Z.~Liu, Y.~Lin, Y.~Cao, H.~Hu, Y.~Wei, Z.~Zhang, S.~Lin, and B.~Guo, ``Swin
  transformer: Hierarchical vision transformer using shifted windows,'' in
  \emph{Proceedings of the IEEE/CVF international conference on computer
  vision}, 2021, pp. 10\,012--10\,022.

\bibitem{touvron2023llama}
H.~Touvron, L.~Martin, K.~Stone, P.~Albert, A.~Almahairi, Y.~Babaei,
  N.~Bashlykov, S.~Batra, P.~Bhargava, S.~Bhosale \emph{et~al.}, ``Llama 2:
  Open foundation and fine-tuned chat models,'' \emph{arXiv preprint
  arXiv:2307.09288}, 2023.

\bibitem{alsentzer2019publicly}
E.~Alsentzer, J.~Murphy, W.~Boag, W.-H. Weng, D.~Jindi, T.~Naumann, and
  M.~McDermott, ``Publicly available clinical bert embeddings,'' in
  \emph{Proceedings of the 2nd clinical natural language processing workshop},
  2019, pp. 72--78.

\bibitem{paszke2019pytorch}
A.~Paszke, S.~Gross, F.~Massa, A.~Lerer, J.~Bradbury, G.~Chanan, T.~Killeen,
  Z.~Lin, N.~Gimelshein, L.~Antiga \emph{et~al.}, ``Pytorch: An imperative
  style, high-performance deep learning library,'' \emph{Advances in neural
  information processing systems}, vol.~32, 2019.

\bibitem{loshchilov2017decoupled}
I.~Loshchilov and F.~Hutter, ``Decoupled weight decay regularization,''
  \emph{arXiv preprint arXiv:1711.05101}, 2017.

\end{thebibliography}
}

\end{document}